\documentclass[reprints,article,accept,pdftex,moreauthors]{Definitions/mdpi} 

\newcommand*\rot{\rotatebox{90}}

\firstpage{1} 
\pubvolume{1}
\issuenum{1}
\articlenumber{0}
\pubyear{2022}
\copyrightyear{2022}
\datereceived{} 
\dateaccepted{} 
\datepublished{} 
\hreflink{https://doi.org/} 

\Title{Empowering Wildlife Guardians: An Equitable Digital Stewardship and Reward System for Biodiversity Conservation using Deep Learning and 3/4G Camera Traps}

\TitleCitation{Title}

\Author{Paul Fergus $^{1}$\orcidA{}, Carl Chalmers $^{1}$, Steven Longmore $^{1}$, Serge Wich $^{1}$, Carmen Warmenhove $^{2}$, Jonathan Swart $^{3}$, Thuto Ngongwane $^{3}$, André Burger $^{3}$, Jonathan Ledgard $^{5}$, and Erik Meijaard $^{4}$}

\AuthorNames{Paul Fergus, Carl Chalmers, Steven Longmore, Serge Wich, Carmen  Warmenhove, Jonathan Swart, Thuto Ngongwane, André Burger, Jonathan Ledgard, and Erik Meijaard }

\AuthorCitation{Fergus, P.; Chalmers, C.; Longmore, S.;  Wich, S.; Warmenhove, C.; Swart, J.; Ngongwane, T.;  Burger, A.; Ledgard, J.; Meijaard, E.;}

\address{%
$^{1}$ \quad Liverpool John Moores University, Byrom Street, Liverpool, United Kingdom; p.fergus@ljmu.ac.uk, c.chalmers@ljmu.ac.uk, s.longmore@ljmu.ac.uk, s.wich@ljmu.ac.uk \\
$^{2}$ \quad Gap Africa Projects, PO Box 198, Chessington, United Kingdom; carmen.warmenhove@gapafricaprojects.com \\
$^{3}$ \quad Welgevonden Game Reserve, PO Box 433, Vaalwater, South Africa; jonathan@welgevonden.org, andre@welgevonden.org \\
$^{4}$ \quad Borneo Futures, PGGMB Building, Jalan Kianggeh, Bandar Seri Begawan BS8111, Brunei Darussalam; emeijaard@borneofutures.org \\
$^{5}$ \quad Artificial Intelligence Centre, Czech Technical University, 166 36 Prague 6, Czechia, eternautical@gmail.com}

\corres{Correspondence: p.fergus@ljmu.ac.uk; Tel.: +44-151-231-2629}

\secondnote{These authors contributed equally to this work.}

\abstract{The biodiversity of our planet is under threat, with approximately one million species expected to become extinct within decades. The reason; negative human actions, which include hunting, overfishing, pollution, and the conversion of land for urbanisation and agricultural purposes. Despite significant investment from charities and governments for activities that benefit nature, global wildlife populations continue to decline. Local wildlife guardians have historically played a critical role in global conservation efforts and have shown their ability to achieve sustainability at various levels. In 2021, COP26 recognised their contributions and pledged US\$1.7 billion per year; however this is a fraction of the global biodiversity budget available (between US\$124 billion and US\$143 billion annually) given they protect 80\% of the planets biodiversity. This paper proposes a radical new solution based on "Interspecies Money," where animals own their own money. Creating a digital twin for each species allows animals to dispense funds to their guardians for the services they provide. For example, a rhinoceros may release a payment to its guardian each time it is detected in a camera trap as long as it remains alive and well. To test the efficacy of this approach 27 camera traps were deployed over a 400km${^2}$ area in Welgevonden Game Reserve in Limpopo Province in South Africa. The motion-triggered camera traps were operational for ten months and, using deep learning, we managed to capture images of 12 distinct animal species. For each species, a makeshift bank account was set up and credited with £100. Each time an animal was captured in a camera and successfully classified, 1 penny (an arbitrary amount - mechanisms still need to be developed to determine the real value of species) was transferred from the animal account to its associated guardian. The trial demonstrated that it is possible to achieve high animal detection accuracy across the 12 species with a sensitivity of 96.38\%, specificity of 99.62\%, precision of 87.14\%, F1 score of 90.33\%, and an accuracy of 99.31\%. The successful detections facilitated the transfer of £185.20 between animals and their associated guardians.}


\begin{document}
	
\section{Introduction}
{O}{ur} planet is a diverse and complex ecosystem that is home to approximately 8.7 million unique species \cite{mora2011many}. The United Nations Sustainable Development Goals report revealed that one million of these species will become extinct within decades \cite{united2019report}. While the situation is critical, the report emphasises that we can still make a difference if we coordinate efforts at a local and global level. Humans have played a major role in every mammal extinction that has occurred over the last 126,000 years \cite{andermann2020past}. This is due to hunting, overharvesting, the introduction of invasive species, pollution, and the conversion of land for crop harvesting and urban construction \cite{pereira2012global}. The illegal wildlife trade, fuelled by the promotion of medicinal myths and the desire for luxury items, has also become a significant contributor to the decline in biodiversity \cite{ellis2013tiger}, \cite{weru2016wildlife}. According to the United Nations Environment Programme, the illegal wildlife trade is estimated to have an annual value of US\$8.5 billion \cite{united2016report}, \cite{gonzalez2022influence}. The white rhinoceros commands the highest price at US\$368,000, with the tiger close behind at US\$350,193 \cite{mcclenachan2016rethinking}. Animal body parts, which are highly sought-after, such as rhinoceros horns, can fetch up to US\$65,000 per kilogram, making them more valuable than gold, heroin, or cocaine \cite{eikelboom2020will}. Pangolins are the most trafficked mammal in the world and although the value of their scales is significantly less than rhinoceros horn (between US\$190/kg and US\$759.15/kg) they are traded by the ton \cite{sharma2020people}. In 2015, 14 tons of pangolin scales, roughly 36,000 pangolins, was seized at a Singapore port with a black market value of US\$39 million \cite{mckirdy2019record}.
\par
Charities, governments and NGOs protect wildlife and their habitats by raising money, developing policies and laws, and lobbying the public to fund conservation projects worldwide \cite{raustiala1997states}. In 2019, the total funding for biodiversity preservation was between US\$124 and US\$143 billion \cite{white2022price}. The funds were split 1\% towards nature-based solutions and carbon markets \cite{girardin2021nature}, 2\% for philanthropy and conservation NGOs \cite{holmes2012biodiversity}, 4\% for green financial products \cite{wang2016role}, 5\% for sustainable supply chains \cite{linton2007sustainable}, 5\% for official development assistance \cite{blunt2011meaning}, 6\% for biodiversity offsets (in agriculture, infrastructure, and extractive industries that unavoidably and negatively impact nature) \cite{bull2013biodiversity}, 20\% for natural infrastructure (such as reefs, forests, wetlands, and other natural systems that provide habitats for wildlife and essential ecosystem services like watershed and coastal protection) \cite{da2017ecosystems}, and finally, 57\% for domestic budgets and tax policy (to direct and influence the economy in ways that increase specific revenue types and discourage activities that harm nature) \cite{pretty2001policy}.
\par
The world's most impoverished individuals, numbering around 720 million, tend to inhabit regions where safeguarding biodiversity is of utmost importance \cite{estrada2022global}, \cite{turner2012global}, \cite{ledgard2022interspecies}. Their cultures, spirituality, and deep-rooted connections to the environment are intertwined with biodiversity \cite{reyes2022recognizing}, and traditionally, conservation practices that involve local people as wildlife guardians have been successful in preventing biodiversity and habitat loss \cite{dawson2021role}, \cite{ruckelshaus2020ipbes}. Yet, they have historically received almost zero economic incentive to protect their surroundings \cite{ledgard2022interspecies}. In 2021, COP26 redressed this issue and pledged US\$1.7 billion annually to local stakeholders in recognition of the biodiversity stewardship services they provide \cite{haenssgen2022implementation}. However, the allocation falls short of the US\$124 billion - US\$143 billion annual global biodiversity budget given that they maintain 80\% of the planets biodiversity \cite{bandiaky2023indigenous}, \cite{lairdconnecting}. Most of the global biodiversity budget is spent in industrialised countries; only a tiny fraction actually ends up in the hands of the extreme poor. In this paper, we propose an innovative solution based on "Interspecies Money," \cite{ledgard2022interspecies} which involves the allocation of funds to animals which they can use to pay local wildlife guardians for the services they provide. Each species group has a digital twin \cite{sharef2022applications} that serves as its identity, and when an animal is detected on camera, it can release funds to its guardian. For example, a giraffe could dispense a few dollars to its guardian each time it is photographed, or they could receive thousands of dollars each time an orangutan is detected, as long as it is alive and well. 
\par
The proposed solution outlined in this paper uses deep learning and 3/4G camera traps to identify animal species and facilitate financial transactions between wildlife accounts and local stakeholders. It differs from the original "Interspecies Money" concept in that it does not use individual identification but simply detections of the species. Using a region-based model, animals are detected in images as and when they are captured in camera traps installed in Welgevonden Game Reserve in Limpopo Province in South Africa. Each animal species group is given a makeshift bank account with £100 credit. Every time an animal is successfully classified in an image, 1 penny is transferred from the animal account to the guardian. Note that this is an arbitrary amount and mechanisms still need to be developed to set the value of species. Figure 1 shows an example detection using our deep learning approach. In this example, three pence would be transferred from the species account to the guardian.
\begin{figure}[htp] 
	\centering
	\includegraphics[width=1\linewidth]{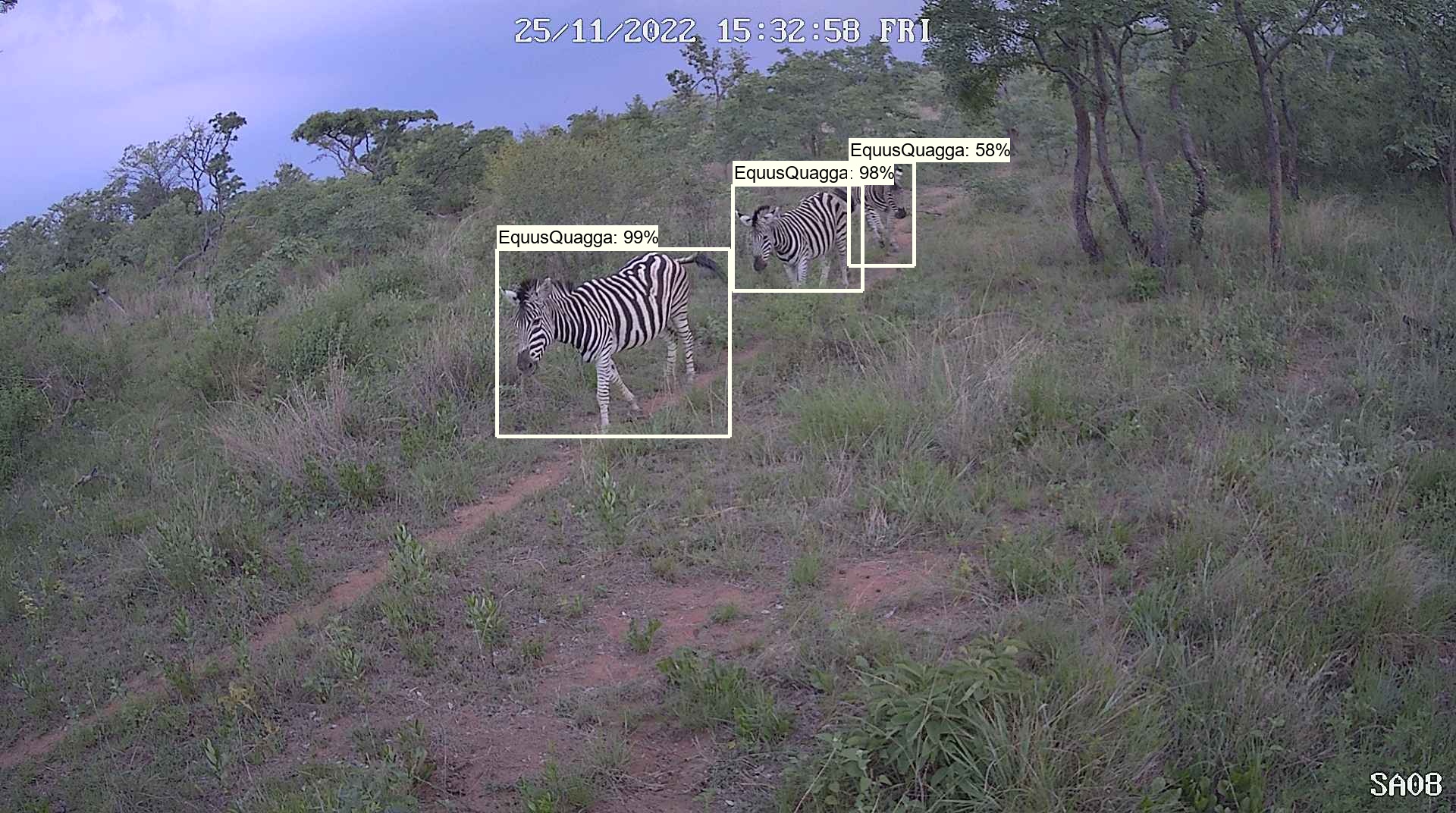}
	\caption{An example detection of a dazzle of \textit{Equus quagga} captured during the trial in Welgevonden Game Reserve in Limpopo in South Africa. Following this detection, 3 pence was transferred from the animal account to the guardian account.}
	\label{fig1} 
\end{figure}
\par
The remainder of this paper delves deeper into the key points discussed in the introduction: Section 2 provides a brief history of conservation, which serves as a foundation for the approach taken in this paper. Section 3 outlines the Materials and Methods used in the trial which posits a new solution for addressing the challenges described. In Section 4 the results are presented before they are discussed in Section 5. Finally, Section 6 concludes the paper and provides suggestions for future work.

\section{A Brief History of Conservation}
Conservation is a multi-dimensional movement that involves political, environmental, and social efforts to manage and protect animals, plants, and natural habitats \cite{escobar1998whose}, \cite{chesson2000mechanisms}. During the "Age of Discovery" in the 15th to 17th century \cite{parry2010age}, sport hunters in the US formed conservation groups to combat the massive loss of wildlife caused by European settlers \cite{10.2307/3984377}. As local policies emerged, people living close to areas where biodiversity was protected, lost property, land, and hunting rights \cite{shaw2021indigenous}. Settlers criminalised poaching, which was associated with local people who often hunted and fished for their survival \cite{hernandez2022fresh}. Widespread laws led to the creation of protected areas and national parks \cite{runte1997national}, established within the context of colonial subjugation \cite{oguamanam2022indigenous}, economic deprivation \cite{cornell1992can}, and systematic oppression of local communities \cite{dominguez2020decolonising}. 
\par
As a result, local people engaged in "illegal" hunting to meet subsistence needs \cite{cooney2018wild}, earn income or status \cite{cooney2020engaging}, pursue traditional practices of cultural significance \cite{lyver2019biocultural}, or address contemporary and historical injustices linked with conservation \cite{cooney2020engaging}. They viewed settlers (including sport hunters) as unwanted interlopers who stole their lands \cite{hernandez2022fresh}. The industrial revolution in the 19th and early 20th centuries \cite{ashton1997industrial}, marked by larger populations and working communities, further escalated the demand for natural resources, resulting in increased biodiversity loss \cite{hawken2013natural, roser2013world}. Today, many local communities in ecologically unique and biodiversity rich regions of the world still perceive conservation as a Western construct created by non-indigenous peoples who continue to exploit their lands and natural resources \cite{schmink2019political}.
\par
For decades, conservationists have debated whether it is human activity or climate change that has driven species extinctions and whether the loss of biodiversity is a recent phenomenon \cite{andermann2020past}. Studies have provided compelling evidence to show that it is in fact humans who are responsible for the wave of extinctions that have occurred since the Late Pleistocene, 126,000 years ago \cite{tilman1994habitat}. For example, toward the end of the Rancholabrean faunal age around 11,000 years ago, a substantial number of large mammals vanished from North America, which included woolly mammoths \cite{nogues2008climate}, giant armadillos \cite{martin2005twilight} and three species of camel \cite{heintzman2015genomic}. Similar extinctions were seen  in New Zealand when the \textit{Dinornithiformes} (Moa) became extinct about 600 years ago \cite{diamond1989present}, \cite{anderson1989mechanics}, and in Madagascar where the \textit{Archaeoindris fontoynontii} (giant lemur) disappeared between 500 and 2000 years ago \cite{perez2005evidence}. Many believe that species and population extinction is a natural phenomena \cite{ceballos2010sixth}, but the evidence suggests that human activity is accelerating species extinction and biodiversity loss \cite{cowie2022sixth}.     
\par
Despite the efforts to protect biodiversity and natural habitats, we are sleepwalking ourselves into a sixth mass extinction \cite{barnosky2011has}. Economic systems driven by limitless growth continue to negatively impact conservation efforts \cite{wiedmann2020scientists}. Rapid development and industrial expansion is depleting natural resources \cite{brown2000can} and intensifying the conversion of large stretches of land for human use \cite{opoku2019biodiversity}. The Earth's forests and oceans are persistently exploited by major corporations who view the planet's natural resources as capital stock \cite{almond2020living}, \cite{welford2013hijacking}. Economic models and financial markets treat natural systems as assets to be used immediately, leading to the abuse of nature for short-term profits with little regard for the long-term costs to society and the environment \cite{helm2015natural}. While The Economics of Ecosystems and Biodiversity (TEEB) attempt to hold large corporations to account \cite{kumar2012economics}, many believe we need nothing short of a redesign of corporations themselves if we are to successfully enable a transition to a ‘Green Economy' \cite{sukhdev2014economics}. Conservationists agree that biodiversity and natural systems are essential for human survival and economic prosperity, but criticise the big corporations and political systems that prioritise immediate economic gains at the expense of the prosperity and well-being of both current and future generations \cite{lubchenco1998entering}. 
\par
The importance of involving local stakeholders as essential contributors in biodiversity monitoring and conservation efforts is emphasized in current perspectives \cite{wells2004integrating}. Recognizing their role as capable natural resource managers, equitable schemes have been introduced to promote their engagement in locally-grounded social-impact assessments that consider the diverse implications of human activities in biodiversity-rich areas \cite{parks2023transforming}. These efforts are largely driven by the Durban Accord led by the International Union for Conservation of Nature (IUCN) \cite{brosius2004indigenous}, which advocates for new governance approaches in protected areas to promote greater equity in local systems \cite{zurba2019indigenous}, \cite{ip2021wwf}. This necessitates a fresh and innovative strategy that upholds conservation objectives while inclusively integrating the interests of all stakeholders involved. This integrated approach aims to foster synergy between conservation, the preservation of life support systems, and sustainable development. While this paper does not claim to comprehensively address all the issues raised, it does offer a rudimentary tool that may help implement such a strategy by quantitatively accounting for biodiversity protection and equitable revenue sharing.
\section{Materials and Methods}
This section describes the implementation details for the digital stewardship and reward system posited in this paper. The section begins with a discussion on the training data collected for a Sub-Saharan Africa deep learning model which is trained to detect 12 different animal species. A second data set, used to evaluate the trained model, is also presented. This is followed by a discussion on the Faster R-CNN architecture \cite{ren2015faster} and its deployment in Conservation AI \cite{conservationai2023}, \cite{chalmers2021modelling} to classify animals and validate the revenue-sharing scheme. The section is concluded with an overview of the performance metrics selected to evaluate the trained model and the inference tasks conducted during the trial.
\subsection{Data Collection and Pre-processing}
The Sub-Saharan Africa model is trained with camera trap images of animals obtained from Conservation AI partners, which include \textit{Equus quagga}, \textit{Giraffa camelopardalis}, \textit{Canis mesomelas}, \textit{Crocuta crocuta}, \textit{Tragelaphus oryx}, \textit{Connochaetes taurinus}, \textit{Acinonyx jubatus}, \textit{Loxodonta africana}, \textit{Hystrix cristata}, \textit{Papio sp}, \textit{Panthera leo} and \textit{Rhinocerotidae} - the 12 species considered in this study. The class distributions can be seen in Figure 2.
\begin{figure}[htp] 
	\centering
	\includegraphics[width=1\linewidth]{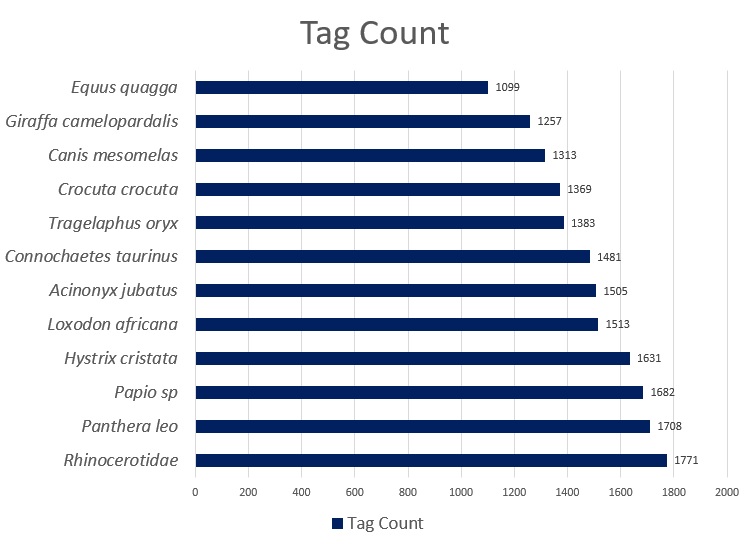}
	\caption{Species Distribution for the Sub-Saharan Training Dataset. The largest number of tags was for the \textit{Rhinocerotidae} (1771) and the lowest was for the \textit{Equus quagga} (1099).}
	\label{fig2} 
\end{figure}
\par
Following quality checks, between 1099 and 1771 tags per species were retained (17,712 in total). Tags are labelled binding boxes that mark the location of an object of interest (animal) within an image. Binding boxes are added to an image using the Conservation AI tagging website which are serialised as coordinates in an XML file using the PASCAL VOC format \cite{lin2014microsoft}. The Conservation AI tagging site is shown in Figure 3. 
\begin{figure}[htp] 
	\centering
	\includegraphics[width=1\linewidth]{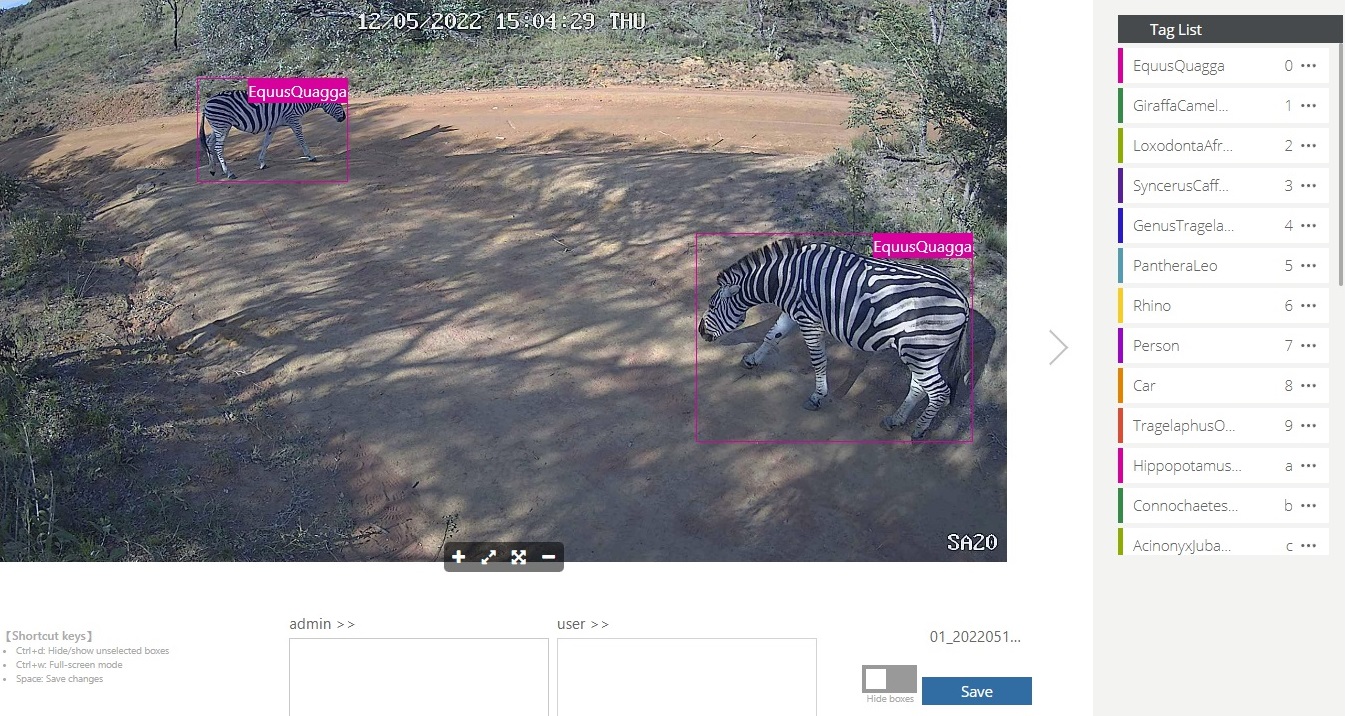}
	\caption{Conservation AI Tagging Site. This example shows two \textit{Equus quagga} tags.}
	\label{fig3} 
\end{figure}
\par
XML files are an intermediary representation used to generate TFRecords (a simple format for storing a sequence of binary records). Before the TFRecords are created, the tagged dataset is randomly split into training and validation sets (90\% - 15,941 tags) and validation (10\% - 1,771). Using the Tensorflow Object Detection API the training and validation datasets are serialised into the two separate TFRecords and used to train the Sub-Saharan African model. The trained model was evaluated over the course of the trial with images obtained from 27 fixed Reolink Go 3/4G cameras installed in Weldgevonden Game Reserve in Limpopo Province in South Africa. Figure 4 shows an example of a camera being installed. 
\begin{figure}[htp] 
	\centering
	\includegraphics[width=1\linewidth]{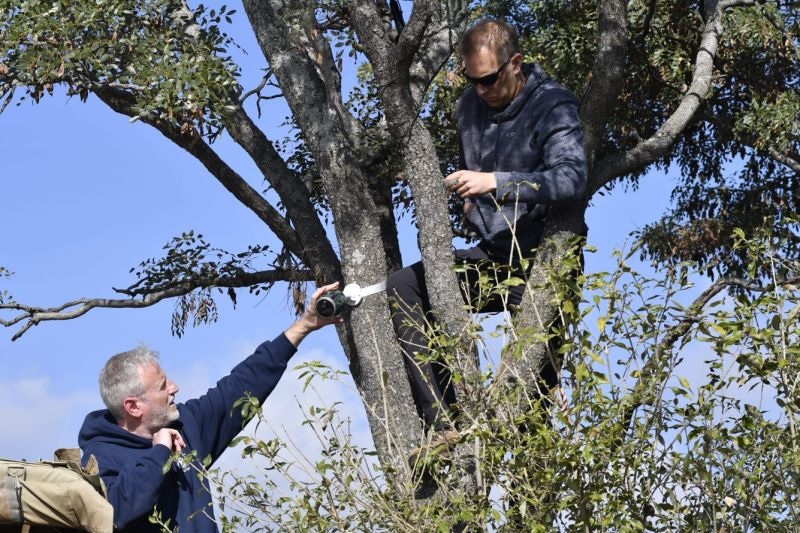}
	\caption{Two of the authors of the paper fitting one of the cameras for the trial in Welgevonden Game Reserve.}
	\label{fig4} 
\end{figure}
\par
The camera resolution was configured to 1920 x 1072 pixels with a Dots Per Inch (dpi) of 96. This configuration closely matches the aspect ratio resizer set in the \textit{pipeline.config} used during the training of the Faster-RCNN. The Infrared (IR) trigger \cite{welbourne2016passive} was set in high sensitivity and each camera was fitted with a camouflage sleeve and fastened to a tree, once a connection to the Vodacom 3/4G network was established (an audible "Connection Succeeded" message is given). Each camera contains a rechargeable lithium-ion battery that is charged using a solar panel fastened to the tree and connected to the camera via a USB Type-B cable. The cameras have an IP65 waterproof rating, providing protection from low pressure rainfall. Designed for security purposes, the cameras have a much wider aspect ratio than conventional camera traps, enabling the detection of animals at farther and wider distances. The installation process spanned seven days and covered an area of 400km${^2}$ (36,000 Hectares). Camera sites were chosen along game paths, water sources, and grazing lawns. The cameras and solar panels were screwed into \textit{Burkea africana} trees and out of reach of \textit{Loxodonta africana} which are known to destroy camera traps in the reserve.
\par
The Reolink Go cameras were donated to us for the study by Reolink and the 3/4G SIM cards were donated to us by Vodacom. The installed camera traps captured 12 different species over a ten month period, which starting in May 2022 and ending in February 2023. During the trial period, 18,520 detections were made and 19,380 blanks reported, totalling 37,900 images. Figure 5 summarises the number of species identified during the trial.
\begin{figure}[htp] 
	\centering
	\includegraphics[width=1\linewidth]{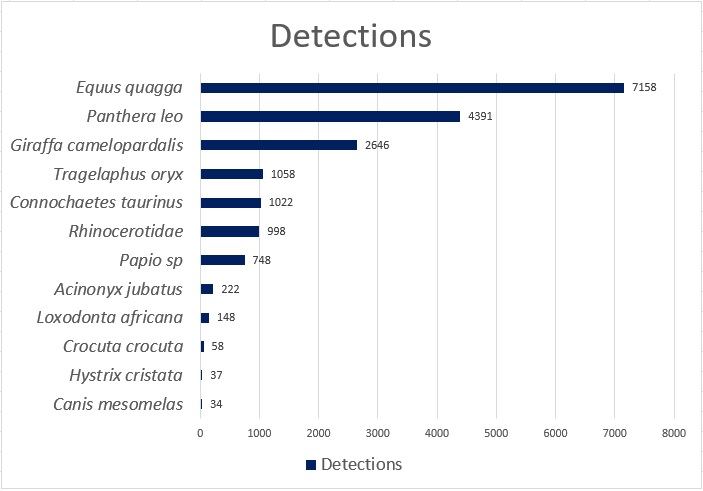}
	\caption{Species Distribution for Detections Captured during the Trial}
	\label{fig5} 
\end{figure}
\par
The 37,900 images were used to generate ten subsets of the data to make the evaluation more achievable. A 5\% margin of error and a confidence level of 95\% in each dataset was maintained using 29 images from each of the 12 animal classes and 29 randomly selected images from the blanks (each of the 10 datasets contained 377 images). The evaluation metrics were calculated for each of the 10 datasets and averaged to produce a final set of metrics for the model.
\subsection{Faster R-CNN}
The Faster R-CNN architecture was trained to detect and classify 12 animal species \cite{ren2015faster}. The architecture comprises three components: a) convolutional neural network (CNN) \cite{gu2018recent} that generates feature maps \cite{ren2016object} and performs classification, b) a region proposal network (RPN) \cite{ren2015faster} that generates Regions of Interest (RoI), and c) a regressor that locates each object in the image and assigns a class label. Figure 6 shows the Faster R-CNN architecture.
\begin{figure}[htp] 
	\centering
	\includegraphics[width=0.60\linewidth]{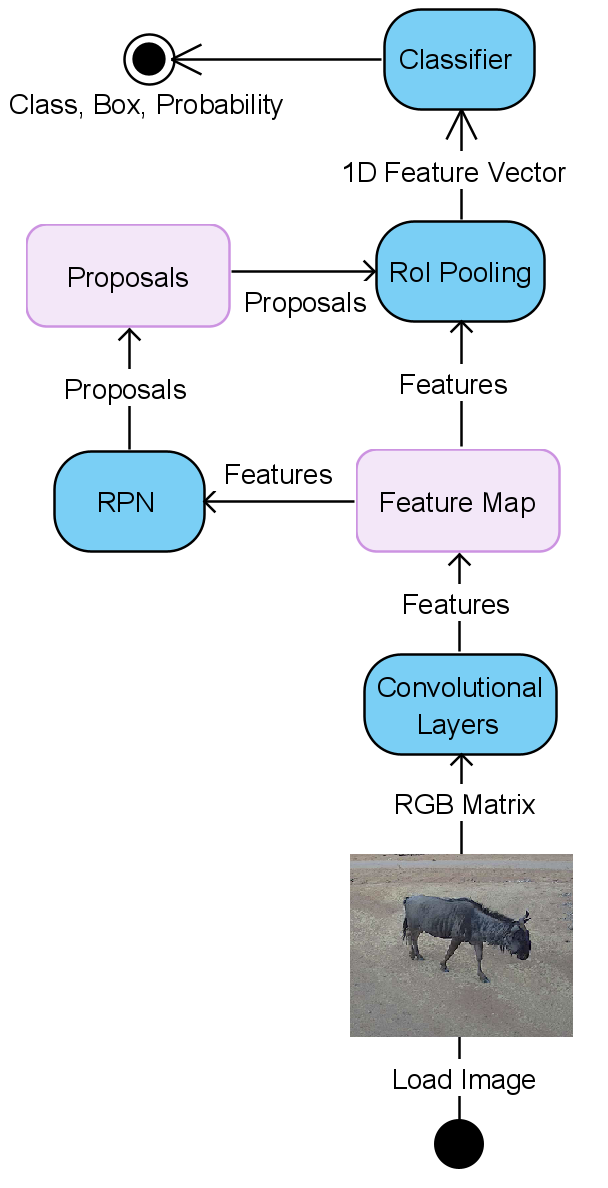}
	\caption{Faster R-CNN architecture showing the base CNN layer, RPN, RoI Pooling and final Fully Connected Classifier.}
	\label{fig6} 
\end{figure}
\par
The RPN is a crucial component in the Sub-Saharan Africa model, as it identifies potential animal species in camera trap images by leveraging the features learned in the base network (ResNet101 in this case \cite{he2016deep}). Unlike early R-CNN networks \cite{girshick2015fast}, which relied on a selective search approach \cite{uijlings2013selective} to generate region proposals at the pixel level, the RPN operates at the feature map level, generating bounding boxes of different sizes and aspect ratios throughout the image, as depicted in Figure 7.
\begin{figure}[htp] 
	\centering
	\includegraphics[width=0.88\linewidth]{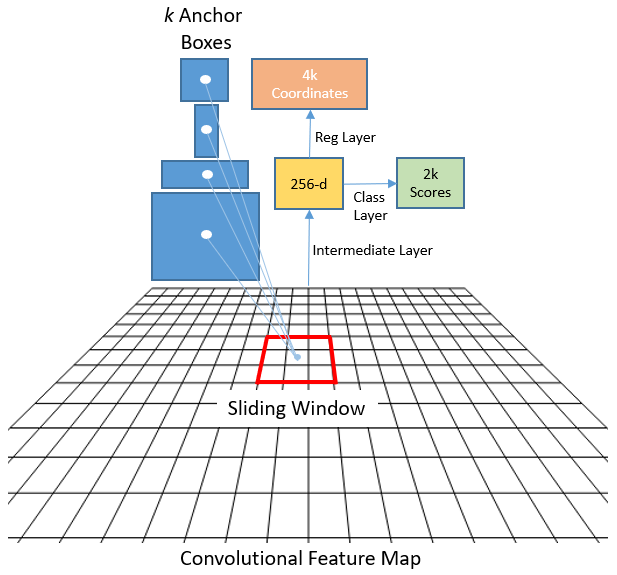}
	\caption{Region Proposal Network showing the different size anchor boxes and the $4k$ Coordinates and $2k$ Scores}
	\label{fig7} 
\end{figure}
\par
The RPN achieves this by employing anchors or fixed bounding boxes, represented by 9 distinct size and aspect ratio configurations, to predict object locations. It is implemented as a CNN, with the feature map supplied by the base network. For each point in the image, a set of anchors is generated, with the feature map dimensions remaining consistent with those of the original image.
\par
The RPN produces two outputs for each anchor bounding box: a probability objectness score and a set of bounding box coordinates. The first output is a binary classification that indicates whether the anchor box contains an object or not, while the second provides a bounding box regression adjustment. During the training process, each anchor is classified as belonging to either a foreground or background category. Foreground anchors are those that have an Intersection over Union (IoU) greater than 0.5 with the ground-truth object, while background anchors are those that do not. The IoU is defined as the ratio of the intersection to the union of the anchor box and the ground truth box. To create mini-batches, 256 balanced foreground and background anchors are randomly sampled, and each batch is used to calculate the classification loss using binary cross-entropy. If there are no foreground anchors in a mini-batch, those with the highest IoU overlap with the ground truth objects are selected as foreground anchors to ensure that the network learns from samples and targets. Additionally, anchors marked as foreground in the mini-batch are used to calculate the regression loss and transform the anchor into the object. The IoU is defined as: 
\begin{equation*} IoU=\frac{Anchor \ box\cap Ground \ Truth \ box}{Anchor \ box\cup Ground \ Truth \ box} \tag{1} \end{equation*} 
\par
Since anchors can overlap, proposals may also overlap on the same object. To address this, Non-Maximum Suppression (NMS) is performed to eliminate intersecting anchor boxes with lower IoU values \cite{neubeck2006efficient}. An IoU greater than 0.7 is indicative of positive object detection, while values less than 0.3 describe background objects. It is important to exercise caution while setting the IoU threshold as setting it too low may lead to missed proposals for objects while setting it too high may result in too many proposals for the same object. Typically, an IoU threshold of 0.6 is sufficient. Once NMS is applied, the top \textit{N} proposals sorted by score are selected.
\par
The loss functions for both the classifier and bounding box calculation are defined as:
\begin{align*} & L_{cls}(p_{i}, p_{i}^{\ast})=- (p_{i}^{\ast}log(p_{i})+(1-p_{i}^{\ast})log(1-p_{i})) \tag{4}\\ & L_{reg}(t_{i},t_{i}^{\ast})= \Sigma_{i\in\{x,y,w,h\}}smooth_{L1}(t_{i}-t_{i}^{\ast}) \tag{5} \end{align*}
where 
\begin{equation*} smooth_{L1}(t_{i}-t_{i}^{\ast})=\begin{cases} 0.5x^{2}& if\vert t_{i}-t_{i}^{\ast}\vert < 1\\ \vert x\vert -0.5& 0ther \end{cases} \tag{6} \end{equation*}
\textit{$p_{i}$} the object possibility, \textit{$t_{i}$} the $4k$ anchor coordinate, \textit{$p_{i}^{*}$} the ground truth label, \textit{$t^{*}$} the ground truth coordinate, $L_{cls}$ the classification loss (log loss), and $L_{reg}$ the regression loss (smooth L1 loss)
\par
After generating object proposals in the RPN step, the next task is to classify and assign a category to each bounding box. In the Faster R-CNN framework, this is accomplished by cropping the convolutional feature map using each proposal and then resizing the crops to 14 x 14 x convdepth using interpolation. To obtain a final 7 x 7 x 512 feature map for each proposal via RoI pooling, max pooling with a 2 x 2 kernel is applied after cropping. These default dimensions are set by the Fast R-CNN \cite{girshick2015fast}, but can be customised depending on the specific use case for the second stage.
\par
The Fast R-CNN architecture takes the 7 x 7 x 512 feature map for each proposal, flattens it into a one-dimensional vector and passes the vector through two fully-connected layers of size 4096 with Rectifier Linear Unit (ReLU) activation \cite{agarap2018deep}. To classify the object category, an additional fully-connected layer is implemented with \textit{N}+1 units, where \textit{N} is the total number of classes and the extra unit corresponds to background objects. Simultaneously, a second fully-connected layer with \textit{4N} units is implemented for predicting the bounding box regression parameters. These 4 parameters are $\Delta_{center_x}$, $\Delta_{center_y}$, $\Delta_{width}$, and $\Delta_{height}$ for each of the \textit{N} possible classes. The Fast R-CNN architecture is illustrated in Figure 8.
\begin{figure}[htp] 
	\centering
	\includegraphics[width=0.88\linewidth]{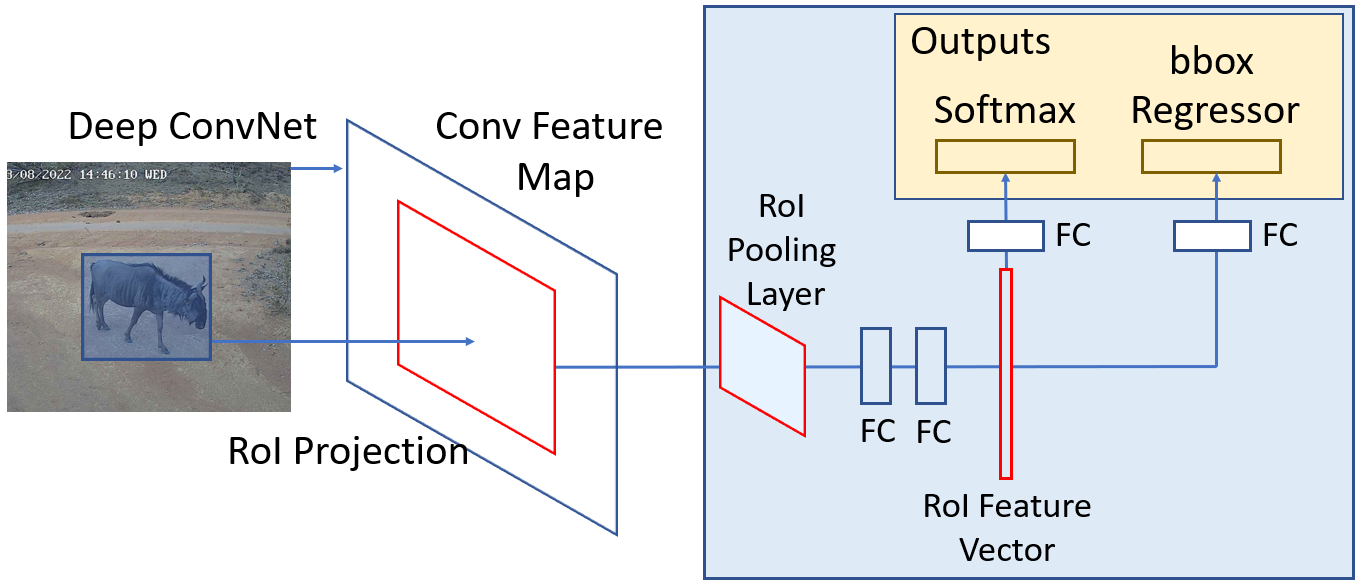}
	\caption{Fast R-CNN architecture shows the RoI Pooling Layer and the two fully connected layers with associated Softmax and Binding Box Regressor outputs.}
	\label{fig8} 
\end{figure}
\par
Targets in the Fast R-CNN are computed in a similar way to RPN targets, but with different classes taken into account. Proposals with an IoU greater than 0.5 with any ground truth box are assigned to that ground truth. Proposals with an IoU between 0.1 and 0.5 are designated as background, while proposals with no intersection are ignored. Targets for bounding box regression are computed for proposals that have been assigned a class based on the IoU threshold by determining the offset between the proposal and its corresponding ground-truth box. The Fast R-CNN is trained using backpropagation \cite{rumelhart1986learning} and Stochastic Gradient Descent \cite{robbins1951stochastic}. The loss function in the Fast R-CNN is calculated as follows:
\begin{equation*} L(p,u,t^{u},\ v)=L_{cls}(p,u)+\lambda.\ [\mathrm{u}\geq 1]L_{reg}(t^{u},\nu) \tag{7} \end{equation*}
where $p$ represents the object probability, $u$ represents the predicted classification class, $t$ represents the ground truth label, and $v$ represents the ground truth coordinates for class $u$. Specifically, the classification loss function $L_{cls}$ is given by:
\begin{equation*} L_{cls}(p,u)=-\log(\frac{e^{p_{u}}}{\Sigma_{j=1}^{K}e^{p_{j}}}) \tag{8} \end{equation*}
where $p$ is the object possibility, $u$ is the classification class, and $K$ is the total number of classes. $L_{reg}$ for bounding box regression can be calculated using the equation described in 5, with $t^u$ and $v$ as input.
\par
To refine the object detection, the Fast R-CNN applies a bounding box adjustment step, which considers the class with the highest probability for each proposal. Proposals assigned to the background class are ignored. Once the final set of objects have been determined, based on the class probabilities, NMS is applied to filter out overlapping boxes. A probability threshold is also set to ensure that only highly confident detections are returned, thereby minimising false positives.
\par
For the complete Faster R-CNN model, there are two losses for the RPN and two for the R-CNN. The four losses are combined through a weighted sum, which can be adjusted to give the classification losses more prominence than the regression losses or to give the R-CNN losses more influence over the RPNs.
\subsection{Transfer Learning}
The Faster R-CNN model is fine-tuned in this study using the dataset containing 12 animal classes; this is known as transfer learning \cite{pan2010survey}. This crucial technique combats overfitting \cite{ying2019overview}, which is a common problem in deep learning when training with limited data. The base model used is ResNet101, a residual neural network \cite{he2016deep} that is pre-trained on the COCO dataset consisting of 330,000 images and 1.5 million object instances \cite{lin2014microsoft}. Residual neural networks employ a highway network architecture \cite{srivastava2015highway}, which enables efficient training in deep neural networks by using skip connections to mitigate the issue of vanishing or exploding gradients.
\subsection{Model Training}
The model training process is performed on a 3U blade server featuring a 24 Core AMD EPYC7352 CPU processor, 512GB RAM, and 8 Nvidia Quadro RTX8000 graphics cards totalling 384GB of GPU memory \cite{keckler2011gpus}. To create the training pipeline, we leverage TensorFlow 2.5 \cite{goldsborough2016tour}, the TensorFlow Object Detection API \cite{huang2017tensorflow}, CUDA 10.2, and CuDNN version 7.6. The TensorFlow configuration file is customised with several hyperparameters to optimise the training process:
\begin{itemize}
	\item Setting the minimum and maximum coefficients for the aspect ratio resizer to 1024x1024 pixels, respectively, to minimise the scaling effect on the data.
	\item Retraining the default feature extractor coefficient to provide a standard 16-pixel stride length, which helps maintain a high-resolution aspect ratio and improve training time.
	\item Setting the batch size coefficient to sixty-four to ensure that the GPU memory limits are not exceeded.
	\item Setting the learning rate to 0.0004 to prevent large variations in response to the error.
\end{itemize}
In order to improve generalisation and to account for variance in the camera trap images the following augmentation settings were used:
\begin{itemize}
	\item $Random\_adjust\_hue$ which adjusts the hue of an image using a random factor.
	\item $Random\_adjust\_contrast$ which adjusts the contrast of an image by a random factor.
	\item $Random\_adjust\_saturation$ which adjusts the saturation of an image by a random factor.
	\item $Random\_square\_crop\_by\_scale$ which was set with a $scale\_min$ of 0.6 and a $scale\_max$ of 1.3.
\end{itemize}
\par
ResNet101 employs the Adam optimiser to minimise the loss function \cite{kingma2014adam}. Unlike optimisers that rely on a single learning rate (alpha) throughout the training process, such as stochastic gradient descent \cite{bottou2012stochastic}, Adam uses the moving averages of the gradients $m_{t}$ and squared gradients $v_{t}$, along with the parameters beta1/beta2, to dynamically adjust the learning rate. Adam is defined as:
\begin{equation} 
	\begin{aligned}
		m_{t} = \beta_{1}m_{t} - 1 + (1- \beta_{1})g_{t} \\
		v_{t} = \beta_{2}v_{t} - 1 + (1 - \beta_{2})g_{t}^{2}
	\end{aligned}
	\tag{9}
\end{equation} 
where $m_{t}$ and $v_{t}$ are estimates of the first and second moment of the gradients. Both $m_{t}$ and $v_{t}$ are initialised with 0's. Biases are corrected by computing the first and second moment estimates:
\begin{equation} 
	\begin{aligned}
		\hat{m}_{t} = \frac{m_{t}}{1-\beta_{1}^{t}} \\
		\hat{v}_{t} = \frac{v_{t}}{1-\beta_{2}^{t}}
	\end{aligned}
	\tag{10}
\end{equation} 
{\parindent0pt 
	Parameters are updated using the Adam update rule:\\	
}
\begin{equation} 
	\begin{aligned}
		\theta_{t+1} =\theta_{t} - \frac{n}{\sqrt{\hat{v}_{t} + \epsilon}}\hat{m}_t\cdot
	\end{aligned}
	\tag{11}
\end{equation}
\par
To overcome the problem of saturation changes around the mid-point of their input, which is common with sigmoid or hyperbolic tangent (tanh) activations \cite{sharma2017activation}, the ReLU activation function is adopted \cite{nair2010rectified}. ReLU is defined as:
\begin{equation} 
	\begin{aligned}
		g(x) = max(0,x)
	\end{aligned}
	\tag{12}
\end{equation}
\subsection{Inference Pipeline}
The Sub-Saharan Africa model is hosted on a Nvidia Triton Inference Server (version 22.08) on a custom-built machine with an Intel Xeon E5-1630v3 CPU, 256GB of RAM, and an Nvidia Tesla T4 GPU \cite{jahanshahi2020gpu}. The real-time cameras transmit images over 3/4G communications every time the IR sensor is triggered. The trigger distance is set to "High" which supports a 9 meter (30 feet) trigger range \cite{caravaggi2017review}. Images are received from cameras using the Simple Mail Transfer Protocol (SMTP) \cite{postel1982simple} and submitted to the Sub-Saharan Africa model using a RestAPI \cite{masse2011rest}. All data is stored in a MySQL database (images are stored in local directories - the MySQL database contains hyperlinks to all images with detections).
\subsection{BioPay}
BioPay is a simple RestAPI service provided by Conservation AI (this is not a public facing service but an experimental module for research purposes only). The service transfers funds between individual species accounts and a guardian account each time a camera is trigged, and an associated animal is detected (see Figures 1, 10, 11 and 12 for sample detections). To ensure efficient fund management, 12 separate makeshift bank accounts, each dedicated to a distinct species, and a central guardian makeshift account were created using the PayPal Sandbox Developer SDKs \cite{paypaldeveloper}. Upon successful classification of an animal, 0.1 GBP is securely transferred from the corresponding species account to the guardian account. Each account was credited with £100 at the start of the trial.        
\subsection{Evaluation Metrics}
RPNLoss/objectiveness, RPNLoss/localisation, BoxClassifierLoss/classification, BoxClassifierLoss/localisation, and TotalLoss are used to evaluate the model during training \cite{padilla2020survey}, \cite{goldsborough2016tour}. RPNLoss/objectiveness evaluates the model's ability to generate bounding boxes and classify background and foreground objects. While, RPNLoss/localisation measures the precision of the RPN's bounding box regressor coordinates for foreground objects, which is to say, how closely each anchor target is to the nearest bounding box. BoxClassifierLoss/classification measures the output layer/final classifier loss for prediction and BoxClassifierLoss/localisation measures the bounding box regressor's performance in terms of localisation. TotalLoss combines all the losses to provide a comprehensive measure of the model's performance.
\par
The validation set during training is evaluated using mAP (mean average precision), which serves as a standard measure for assessing object detection models. mAP is defined as:
\begin{equation} 
	\begin{aligned}
		mAP = \frac{\sum_{q=1}^{Q} AveP(q)}{Q}
	\end{aligned}
	\tag{13}
\end{equation}
where Q is the number of queries in the set and $AveP(q)$ is the average precision ($AP$) for a given query $a$.
\par
mAP is computed for the binding box locations using the final two checkpoints. The calculation involves measuring the percentage IoU between the predicted bounding box and the ground truth bounding box and is expressed as:
\begin{equation} 
	\begin{aligned}
		IoU = \frac{Area of Overlap}{Area of Union}
	\end{aligned}
	\tag{14}
\end{equation}
\par
The detection accuracy and localisation accuracy are measured using two distinct IoU thresholds, namely @.50 and @.75, respectively. The @.50 threshold evaluates the overall detection accuracy and the higher @.75 threshold focuses on the model's ability to accurately localise objects.
\par
Accuracy, Precision, Sensitivity, Specificity, and F1-Score, are used to evaluate the performance of the trained model during inference, in other words, using the image data collected during the trial. Accuracy is defined as:
\begin{equation}
	\begin{aligned}
		Accuracy = \frac{TP + TN}{TP + FP + TN + FN}
	\end{aligned}
	\tag{15}
\end{equation}
\par
where TP is True Positives, TN is True Negatives, FP is False Positives and FN is False Negatives. The accuracy metric provides an overall assessment of the object detection models ability to inference on unseen data. This metric is often interpreted alongside the other metrics defined below.
\par
Precision is used to assess the models ability to predict true positive detections and is defined as:
\begin{equation}
	\begin{aligned}
		Precision = \frac{TP}{TP + FP}
	\end{aligned}
	\tag{16}
\end{equation}
It measures the fraction of true positive detections out of all detections made by the trained model. In object detection, a true positive detection occurs when the model correctly identifies an object and predicts its location in the image. A high precision indicates that the model has a low rate of false positives, meaning that when it makes a positive detection, it is highly likely to be correct. 
\par
Sensitivity, also known as recall, measures the proportion of true positives correctly identified by the trained model during inference. In other words, it measures the model's ability to detect all positive instances in the dataset. A high sensitivity indicates that the model has a low rate of false negatives, meaning that when an object is present in the image, the model is highly likely to detect it. This metric is defined as:
\begin{equation} 
	\begin{aligned}
		Sensitivity = \frac{TP}{TP + FN}
	\end{aligned}
	\tag{17}
\end{equation}
\par
Specificity measures the proportion of true negative detections that are correctly identified by the model. In object detection, true negative detections refer to areas of an image where there is no object of interest. In this paper we evaluate blank images to satisfy this metric as it is important to ensure that classifications are based on features extracted from animals and not the background. Specificity is defined as:
\begin{equation} 
	\begin{aligned}
		Specificity = \frac{TN}{TN + FP}
	\end{aligned}
	\tag{18}
\end{equation}
\par
Finally, F1-Score combines precision and recall into a single score. A high F1-Score indicates that the model has both high precision and high recall, meaning that it can accurately identify and localise objects in the image. F1-Score is defined as:
\begin{equation} 
	\begin{aligned}
		F1 Score =2 * \frac{Precision * Sensitivity(Recall)}{Precision + Sensivitity(Recall)}
	\end{aligned}
	\tag{19}
\end{equation}
\par
The ground truths for the 10 subsampled datasets where provided by conservationists and biologists who appear as co-authors in the paper. The ground truths are used to calculate the detections generated by the in-trial model.    
\section{Evaluation}
The results obtained during the training of the Sub-Saharan Africa model are presented first. The model is then evaluated in a real world setting to assesses its ability to classify animal species in images captured during the trial. This section is concluded with the results obtained for the financial transactions taken between animals and the guardian bank account when positive detections are made.
\subsection{Training Results for the Sub-Saharan Model}
In the first evaluation, the training set (containing 17712 images - See Figure 2 for species tag distributions) is used to fit the model. The dataset is randomly split, as previously discussed in Section 3.1 and trained over 28000 steps (438 epochs) using a batch size of 64.
\subsubsection{RPN and Box Classification Results for the Training Dataset}
The outcomes depicted in Table 1 indicate that the model is capable of detecting candidate regions of interest with sufficient accuracy (loss=0.0366). The RPN can effectively model localisation (loss=0.0112). Classification loss (loss=0.1833) is higher than previous losses which shows the model is less precise at classifying objects of interest than locating them. Box classifier localisation (loss=0.0261) is comparable to the RPN results and confirms that the model can sufficiently identify candidate bounding boxes. The total loss value (0.2244), combines both the RPN and Box classification losses, and indicates that the model's predictions are relatively close to the ground-truth labels. In most cases, a total loss=0.2244 in object detection is considered good outcome.
\begin{table}[H] 
	\caption{RPN and Box Classification Results for Training.\label{tab5}}
	\newcolumntype{C}{>{\centering\arraybackslash}X}
	\begin{tabularx}{\textwidth}{CCCC}
		\toprule
		\textbf{Metric}	& \textbf{Value}	& \textbf{Steps} & \textbf{Support}\\
		\midrule
		RPNLoss/Objectness				 & 0.0366 & 28k & 15941\\
		RPNLoss/Localisation			 & 0.0112 & 28k & 15941\\
		BoxClassifierLoss/Classification & 0.1833 & 28k & 15941\\
		BoxClassifierLoss/Localisation 	 & 0.0261 & 28k & 15941\\
		Total Loss						 & 0.2244 & 28k & 15941\\
		\bottomrule
	\end{tabularx}
\end{table}
\subsubsection{RPN and Box Classification Results for the Validation Dataset}
Table 2 provides the results for the validation set, which are generally consistent with those produced by the training set. The loss for regions of interest (loss=0.0530) is marginally higher but not significantly. The same applies to RPN localisation (loss=0.0384), which is also slightly higher but not significantly. Both the losses for classification and box classifier localisation, 0.1533 and 0.0242 respectively, are  similar to those in Table 1. The combined losses indicate that the validation set produces good results, as shown in total loss (loss=0.2690) in Table 2. It is worth noting that the training and validation metrics are close, meaning there was no evidence of overfitting during model training.
\begin{table}[H] 
	\caption{RPN and Box Classification Results for Validation.\label{tab5}}
	\newcolumntype{C}{>{\centering\arraybackslash}X}
	\begin{tabularx}{\textwidth}{CCCC}
		\toprule
		\textbf{Metric}	& \textbf{Value}	& \textbf{Steps} & \textbf{Support}\\
		\midrule
		RPNLoss/Objectness					& 0.0530 & 28k & 1771\\
		RPNLoss/Localisation			 	& 0.0384 & 28k & 1771\\
		BoxClassifierLoss/Classification 	& 0.1533 & 28k & 1771\\
		BoxClassifierLoss/Localisation 	 	& 0.0242 & 28k & 1771\\
		Total Loss						 	& 0.2690 & 28k & 1771\\
		\bottomrule
	\end{tabularx}
\end{table}
\subsubsection{Precision and Recall Results for Validation Dataset}
The mAP values in Table 3 provide the mean of the average precisions achieved for all classes using IoU thresholds between .5 and .95 with .05 increments. The result (mAP = 0.7542) indicates that 24.8\% false positives were observed across all classes and IoU thresholds. An mAP of 0.7542 would generally indicate that the model is able to correctly detect objects with a high level of accuracy. According to the precision metrics for objects of different sizes, the model appears to be more proficient at detecting larger-sized objects (0.7815) within images, as opposed to medium and smaller objects (0.3528 and 0.1362 respectively). An mAP of 0.9449 and 0.8601 at IoU thresholds of 0.50 and 0.75 respectively indicate that the model is able to accurately detect objects with a high level of precision across a wide range of IoU thresholds.
\begin{table}[H] 
	\caption{Precision Results for Sub-Saharan Africa Model.\label{tab5}}
	\newcolumntype{C}{>{\centering\arraybackslash}X}
	\begin{tabularx}{\textwidth}{CCCC}
		\toprule
		\textbf{Metric}	& \textbf{Value}	& \textbf{Steps} & \textbf{Support}\\
		\midrule
		Precision/mAP	 	 	& 0.7542 & 28k & 3134\\
		Precision/mAP(Large)	& 0.7815 & 28k & 3134\\
		Precision/mAP(Medium) 	& 0.3528 & 28k & 3134\\
		Precision/mAP(Small)  	& 0.1362 & 28k & 3134\\
		Precision/mAP@.50IOU	& 0.9449 & 28k & 3134\\
		Precision/mAP@.75IOU	& 0.8601 & 28k & 3134\\
		\bottomrule
	\end{tabularx}
\end{table}
\par
The recall values in Table 4 indicate that the model can retrieve 62.39\% of all images among the top 1 images retrieved (Recall/AR@1). As the number of returned images increases, the number of relevant images increases (Recall/AR@10=0.8060 and Recall/AR@100=0.8140). In object detection these are again satisfactory results. The recall values for AR@100 (small, medium, and large) show that the model is better at detecting large and medium objects in images (0.8496 and 0.5079 respectively) opposed to smaller objects (0.3344).
\begin{table}[H] 
	\caption{Recall Results for Sub-Saharan Africa Model.\label{tab5}}
	\newcolumntype{C}{>{\centering\arraybackslash}X}
	\begin{tabularx}{\textwidth}{CCCC}
		\toprule
		\textbf{Metric}	& \textbf{Value}	& \textbf{Steps} & \textbf{Support}\\
		\midrule
		Recall/AR@1		 	  & 0.6239 & 25k & 3134\\
		Recall/AR@10		  & 0.8060 & 25k & 3134\\
		Recall/AR@100 		  & 0.8140 & 25k & 3134\\
		Recall/AR@100(Large)  & 0.8357 & 25k & 3134\\
		Recall/AR@100(Medium) & 0.5079 & 25k & 3134\\
		Recall/AR@100(Small)  & 0.3344 & 25k & 3134\\
		\bottomrule
	\end{tabularx}
\end{table}
\subsection{Trial Results Using the Sub-Saharan African Model}
The trained model was deployed and used in the Welgevonden Game Reserve in Limpopo Province in South Africa to detect 12 animal species captured in camera traps. During the 10 month trial 18,520 images with detections were recorded and there were 19,380 images with no animal in them (blanks). In total 37,800 images were collected. Figure 5 shows the distribution of all animal detections. Due to the size of the dataset, 10 subsets of the data were created to make the evaluation achievable.
\subsubsection{Performance Metrics for Inference}
The results in Table 5 show that the model achieves high accuracy scores for all animal classes, with values between 98.70\% and 100.00\%. The model's precision scores are more varied with scores between 60.00\% and 100.00\%, with the \textit{Acinonyx jubatus} and \textit{Panthera leo} classes having the lowest values. Overall, the average model precision (87.14\%) is considered a good result in object detection. However, there are issues with several classes (\textit{Hystrix cristata}, \textit{Acinonyx jubatus} and \textit{Panthera leo}) which require further investigation. Sensitivity is consistently high for all classes indicating that the model correctly identifies positive samples. Again, the overall sensitivity for the model (96.38\%) is high which shows that a large proportion of actual positive cases are correctly identified by the model. The classes with the lowest sensitivity values (\textit{Tragelaphus oryx} and \textit{Panthera leo} 90.90\% and 90\% respectively) may need further investigation. This seems to be because the model heavily reliant on the colour and shape of the \textit{Tragelaphus oryx}. Specificity is consistently high across all classes. And again, the overall specificity for the model is high (99.62\%) which indicates the model can effectively detect negative cases (i.e., distinguish between blank images and animal classes). Finally, the F1-scores provide the harmonic mean for precision and recall (sensitivity), and most classes are generally high. The overall model F1-Score (90.33\%) is good although addressing issues around precision for the classes mentioned above will improve the F1-score.
\begin{table}[H]
	\caption{Performance Metrics for Inference.\label{tab2}}
	\begin{adjustwidth}{-\extralength}{0cm}
		\newcolumntype{C}{>{\centering\arraybackslash}X}
		\begin{tabularx}{\linewidth}{p{3cm}CCCCCCCCCCC}
			\toprule
			  & \textbf{Fold1} & \textbf{Fold2} & \textbf{Fold3} & \textbf{Fold4} & \textbf{Fold5} & \textbf{Fold6} & \textbf{Fold7} & \textbf{Fold8} & \textbf{Fold9} & \textbf{Fold10} & \textbf{Avg}\\
			\midrule
			\textbf{\textit{Canis mesomelas}}	& & & & & & & & & & & \\										
			Accuracy		& 99.40	& 99.69	& 99.68	& 99.68	& 99.69	& 99.67	& 99.38	& 99.69	& 99.73	& 99.70	& 99.63 \\
			Precision		& 85.71 & 85.71 & 85.71 & 85.71 & 83.33 & 80.00 & 87.50 & 85.71 & 83.33 & 88.89 & 85.16 \\
			Sensitivity		& 85.71 & 100.00& 100.00& 100.00& 100.00& 100.00& 87.50 & 100.00& 100.00& 100.00& 97.32 \\
			Specificity		& 99.69 & 99.68 & 99.67 & 99.68 & 99.68 & 99.67 & 99.68 & 99.69 & 99.72 & 99.69 & 99.69 \\
			F1-Score		& 85.71 & 92.31 & 92.31 & 92.31 & 90.91 & 88.89 & 87.50 & 92.31 & 90.91 & 94.12 & 90.73 \\
			\textbf{\textit{Hystrix cristata}} & & & & & & & & & & & \\											
			Accuracy		& 98.80 & 99.37 & 99.30 & 99.68 & 100.00& 99.35 & 99.38 & 99.38 & 99.73 & 99.70 & 99.48 \\
			Precision		& 63.64 & 77.78 & 66.67 & 88.89 & 100.00& 81.82 & 60.00 & 71.43 & 75.00 & 85.71 & 77.09 \\
			Sensitivity		& 100.00& 100.00& 100.00& 100.00& 100.00& 100.00& 100.00& 100.00& 100.00& 100.00& 100.00 \\
			Specificity		& 98.77 & 99.36 & 99.35 & 99.67 & 100.00& 99.33 & 99.38 & 99.38 & 99.73 & 99.69 & 99.47 \\
			F1-Score		& 77.78 & 87.50 & 80.00 & 94.12 & 100.00& 90.00 & 75.00 & 83.33 & 85.71 & 92.31 & 86.58 \\
			\textbf{\textit{Crocuta crocuta}}	& & & & & & & & & & & \\										
			Accuracy		& 99.40 & 99.69 & 99.36 & 99.68 & 100.00& 100.00& 99.69 & 99.38 & 100.00& 100.00& 99.72 \\
			Precision		& 80.00 & 80.00 & 71.43 & 80.00 & 100.00& 100.00& 80.00 & 71.43 & 100.00& 100.00& 86.29 \\
			Sensitivity		& 80.00 & 100.00& 100.00& 100.00& 100.00& 100.00& 100.00& 100.00& 100.00& 100.00& 98.00 \\
			Specificity		& 99.69 & 99.68 & 99.35 & 99.68 & 100.00& 100.00& 99.69 & 99.38 & 100.00& 100.00& 99.75 \\
			F1-Score		& 80.00 & 88.89 & 83.33 & 88.89 & 100.00& 100.00& 88.89 & 83.33 & 100.00& 100.00& 91.33 \\
			\textbf{\textit{Loxodonta africana}} & & & & & & & & & & & \\											
			Accuracy		& 99.10 & 99.37 & 99.68 & 99.37 & 99.37 & 98.70 & 98.77 & 98.18 & 99.19 & 99.10 & 99.08 \\
			Precision		& 86.96 & 81.82 & 94.44 & 94.44 & 93.75 & 76.47 & 78.95 & 75.00 & 88.89 & 88.24 & 85.90 \\
			Sensitivity		& 100.00& 100.00& 100.00& 94.44 & 93.75 & 100.00& 100.00& 85.71 & 94.12 & 93.75 & 96.18 \\
			Speciality		& 99.04 & 99.35 & 99.66 & 99.66 & 99.67 & 98.64 & 98.71 & 98.73 & 99.43 & 99.37 & 99.23 \\
			F1-Score		& 93.02 & 90.00 & 97.14 & 94.44 & 93.75 & 86.67 & 88.24 & 80.00 & 91.43 & 90.91 & 90.56 \\
			\textbf{\textit{Acinonyx jubatus}} & & & & & & & & & & & \\											
			Accuracy		& 99.10 & 99.69 & 99.68 & 100.00& 99.69 & 99.35 & 99.08 & 99.38 & 99.46 & 99.70 & 99.51 \\
			Precision		& 40.00 & 66.67 & 66.67 & 100.00& 50.00 & 50.00 & 40.00 & 50.00 & 50.00 & 80.00 & 59.33 \\
			Sensitivity		& 100.00& 100.00& 100.00& 100.00& 100.00& 100.00& 100.00& 100.00& 100.00& 100.00& 100.00 \\
			Specificity		& 99.09 & 99.68 & 99.68 & 100.00& 99.68 & 99.34 & 99.07 & 99.38 & 99.45 & 99.70 & 99.51 \\
			F1-Score		& 57.14 & 80.00 & 80.00 & 100.00& 66.67 & 66.67 & 57.14 & 66.67 & 66.67 & 88.89 & 72.98 \\
			\textbf{\textit{Papio sp}}		& & & & & & & & & & & \\									
			Accuracy		& 100.00& 99.69 & 99.68 & 100.00& 99.06 & 99.67 & 99.69 & 97.88 & 99.19 & 99.40 & 99.43 \\
			Precision		& 100.00& 100.00& 100.00& 100.00& 93.33 & 100.00& 100.00& 75.00 & 90.91 & 95.45 & 95.47 \\
			Sensitivity		& 100.00& 95.24 & 94.12 & 100.00& 96.55 & 95.83 & 96.67 & 88.24 & 95.24 & 95.45 & 95.73 \\
			Specificity		& 100.00& 100.00& 100.00& 100.00& 99.31 & 100.00& 100.00& 98.40 & 99.43 & 99.68 & 99.68 \\
			F1-Score		& 100.00& 97.56 & 96.97 & 100.00& 94.92 & 97.87 & 98.31 & 81.08 & 93.02 & 95.45 & 95.52 \\
			\textbf{\textit{Blank}}			& & & & & & & & & & & \\								
			Accuracy		& 100.00& 100.00& 100.00& 100.00& 100.00& 100.00& 100.00& 100.00& 100.00& 100.00& 100.00 \\
			Precision		& 100.00& 100.00& 100.00& 100.00& 100.00& 100.00& 100.00& 100.00& 100.00& 100.00& 100.00 \\
			Sensitivity		& 100.00& 100.00& 100.00& 100.00& 100.00& 100.00& 100.00& 100.00& 100.00& 100.00& 100.00 \\
			Specificity		& 100.00& 100.00& 100.00& 100.00& 100.00& 100.00& 100.00& 100.00& 100.00& 100.00& 100.00 \\
			F1-Score		& 100.00& 100.00& 100.00& 100.00& 100.00& 100.00& 100.00& 100.00& 100.00& 100.00& 100.00 \\
			\textbf{\textit{Rhinocerotidae}}	& & & & & & & & & & & \\										
			Accuracy		& 98.80 & 98.45	& 99.05	& 99.37	& 99.06	& 99.67	& 99.38	& 99.38	& 98.92	& 99.40	& 99.15 \\
			Precision		& 100.00& 97.50 & 93.10 & 97.56 & 90.32 & 100.00& 100.00& 97.37 & 89.19 & 94.44 & 95.95 \\
			Sensitivity		& 91.30 & 90.70 & 96.43 & 97.56 & 100.00& 97.14 & 92.00 & 97.37 &100.00 & 100.00& 96.25 \\
			Specificity		& 100.00& 99.64 & 99.30 & 99.64 & 98.97 & 100.00& 100.00& 99.65 & 98.81 & 99.33 & 99.53 \\
			F1-Score		& 95.45 & 93.98 & 94.74 & 97.56 & 94.92 & 98.55 & 95.83 & 97.37 & 94.29 & 97.14 & 95.98 \\
			\textbf{\textit{Connochaetes taurinus}} & & & & & & & & & & & \\											
			Accuracy		& 97.63 & 99.37 & 98.11 & 98.74 & 98.45 & 98.06 & 98.77 & 97.88 & 97.08 & 98.51 & 98.26 \\
			Precision		& 100.00& 100.00& 93.65 & 100.00& 98.31 & 98.44 & 98.48 & 100.00& 97.47 & 96.49 & 98.28 \\
			Sensitivity		& 89.47 & 96.08 & 96.72 & 92.98 & 93.55 & 92.65 & 95.59 & 90.67 & 89.53 & 94.83 & 93.21 \\
			Specificity		& 100.00& 100.00& 98.44 & 100.00& 99.62 & 99.59 & 99.61 & 100.00& 99.31 & 99.28 & 99.59 \\
			F1-Score		& 94.44 & 98.00 & 95.16 & 96.36 & 95.87 & 95.45 & 97.01 & 95.10 & 93.33 & 95.65 & 95.64 \\
\bottomrule
\end{tabularx}
\end{adjustwidth}
\end{table}

\begin{table}[H]\ContinuedFloat
	\caption{{\em Cont.}\label{tab2}}
	\begin{adjustwidth}{-\extralength}{0cm}
		\newcolumntype{C}{>{\centering\arraybackslash}X}
		\begin{tabularx}{\linewidth}{p{3.5cm}CCCCCCCCCCC}
			\toprule
			  & \textbf{Fold1} & \textbf{Fold2} & \textbf{Fold3} & \textbf{Fold4} & \textbf{Fold5} & \textbf{Fold6} & \textbf{Fold7} & \textbf{Fold8} & \textbf{Fold9} & \textbf{Fold10} & \textbf{Avg}\\
			\midrule
			\textbf{\textit{Tragelaphus oryx}} & & & & & & & & & & & \\											
			Accuracy	& 98.80	& 98.14	& 97.50	& 99.05	& 99.37	& 99.02 & 99.38 & 99.38	& 98.65 & 98.51 & 98.78\\
			Precision	& 89.19 & 97.50 & 100.00& 94.74 & 100.00& 100.00& 100.00& 96.88 & 97.22 & 100.00& 97.55\\
			Sensitivity	& 100.00& 88.64 & 75.00 & 97.30 & 94.29 & 90.00 & 94.44 & 96.88 & 89.74 & 82.76 & 90.90\\
			Specificity	& 98.67 & 99.64 & 100.00& 99.28 & 100.00& 100.00& 100.00& 99.66 & 99.70 & 100.00& 99.69\\
			F1-Score	& 94.29 & 92.86 & 85.71 & 96.00 & 97.06 & 94.74 & 97.14 & 96.88 & 93.33 & 90.57 & 93.86\\
			\textbf{\textit{Giraffa camelopardalis}} & & & & & & & & & & & \\											
			Accuracy	& 99.70 & 100.00& 99.68 & 99.68 & 100.00& 99.67 & 99.69 & 99.69 & 100.00& 100.00& 99.81\\
			Precision	& 100.00& 100.00& 96.55 & 96.00 & 100.00& 96.00 & 96.67 & 100.00& 100.00& 100.00& 98.52\\
			Sensitivity	& 96.67 & 100.00& 100.00& 100.00& 100.00& 100.00& 100.00& 96.15 & 100.00& 100.00& 99.28\\
			Specificity	& 100.00& 100.00& 99.65 & 99.66 & 100.00& 99.64 & 99.66 & 100.00& 100.00& 100.00& 99.86\\
			F1-Score	& 98.31 & 100.00& 98.25 & 97.96 & 100.00& 97.96 & 98.31 & 98.04 & 100.00& 100.00& 98.88\\
			\textbf{\textit{Panthera leo}} & & & & & & & & & & & \\											
			Accuracy	& 97.92 & 98.77 & 99.05 & 99.37 & 100.00& 99.67 & 100.00& 99.69 & 99.46 & 99.40 & 99.33\\
			Precision	& 22.22 & 50.00 & 40.00 & 66.67 & 100.00& 50.00 & 100.00& 66.67 & 33.33 & 60.00 & 53.89\\
			Sensitivity	& 100.00& 100.00& 100.00& 100.00& 100.00& 100.00& 100.00& 100.00& 100.00& 100.00& 90.00\\
			Specificity	& 97.90 & 98.75 & 99.04 & 99.36 & 100.00& 99.67 & 100.00& 99.69 & 99.46 & 99.39 & 99.33\\
			F1-Score	& 36.36 & 66.67 & 57.14 & 80.00 & 100.00& 66.67 & 100.00& 80.00 & 50.00 & 75.00 & 64.52\\
			\textbf{\textit{Equus quagga}} & & & & & & & & & & & \\											
			Accuracy	& 98.80 & 99.69 & 98.42 & 99.05 & 99.69 & 97.44 & 98.77 & 97.88 & 98.92 & 99.40 & 98.80\\
			Precision	& 100.00& 100.00& 100.00& 100.00& 100.00& 96.05 & 100.00& 98.85 & 99.10 & 100.00& 99.40\\
			Sensitivity	& 94.94 & 98.80 & 94.90 & 95.95 & 98.85 & 93.59 & 95.24 & 93.48 & 97.35 & 98.00 & 96.11\\
			Specificity	& 100.00& 100.00& 100.00& 100.00& 100.00& 98.72 & 100.00& 99.58 & 99.61 & 100.00& 99.79\\
			F1-Score	& 97.40 & 99.39 & 97.38 & 97.93 & 99.42 & 94.81 & 97.56 & 96.09 & 98.21 & 98.99 & 97.72\\
			\textbf{Overall Model} & & & & & & & & & & & \\											
			Accuracy	& 99.03 & 99.38 & 99.17 & 99.51 & 99.57 & 99.25 & 99.38 & 99.06 & 99.25 & 99.45 & 99.31\\
			Precision	& 82.13 & 83.61 & 85.25 & 92.62 & 93.00 & 86.83 & 87.82 & 83.72 & 84.96 & 91.48 & 87.14\\
			Sensitivity	& 95.24 & 89.96 & 96.71 & 98.33 & 98.23 & 97.63 & 97.03 & 96.04 & 97.38 & 97.29 & 96.38\\
			Specificity	& 99.45 & 99.68 & 99.55 & 99.74 & 99.76 & 99.58 & 99.68 & 99.50 & 99.59 & 99.70 & 99.62\\
			F1-Score	& 85.38 & 86.19 & 89.09 & 95.04 & 94.88 & 90.64 & 90.84 & 88.48 & 88.99 & 93.77 & 90.33\\
			\bottomrule
		\end{tabularx}
	\end{adjustwidth}
	\noindent{\footnotesize{\textsuperscript{1} This is a table footnote.}}
\end{table}
\subsubsection{Confusion Matrix}
Examining the confusion matrix in Table 6, the results align with those in Table 5. The model accurately predicted all samples for the \textit{Canis mesomelas}, \textit{Hystrix cristata}, \textit{Acinonyx jubatus}, and \textit{Panthera leo} classes. Additionally, the model performed remarkably well for the \textit{Loxodonta africana}, \textit{Papio sp}, and \textit{Blank} classes, with only a few miss-classifications. However, the model miss-classified 42 samples from the \textit{Connochaetes taurinus} class as \textit{Tragelaphus oryx} and 17 samples as \textit{Loxodonta africana}. Similarly, for the \textit{Tragelaphus oryx} class, the model misclassified 15 samples as \textit{Equus quagga} and 11 samples as \textit{Connochaetes taurinus}. This information highlights the specific areas where future re-training is required.
\begin{table}[H] 
	\caption{Confusion Matrix for Sub-Saharan Model Inference.\label{tab5}}
	\newcolumntype{C}{>{\centering\arraybackslash}X}
	\begin{tabularx}{\textwidth}{p{3.5cm}CCCCCCCCCCCCC}
		\toprule
		& \rot{\textit{Canis mesomelas}}	& \rot{\textit{Hystrix cristata}} & \rot{\textit{Crocuta crocuta}} & \rot{\textit{Loxodonta africana}} & \rot{\textit{Acinonyx jubatus}} & \rot{\textit{Papio sp}} & \rot{\textit{Blank}} & \rot{\textit{Rhinocerotidae}} & \rot{\textit{Connochaetes taurinus}} & \rot{\textit{Tragelaphus oryx}} & \rot{\textit{Giraffa camelopardalis}} & \rot{\textit{Panthera leo}} & \rot{\textit{Equus quagga}}\\
		\midrule
		\textit{Canis mesomelas}			& 59	& 0		& 0		& 0		& 2		& 0		& 0		& 0		& 0		& 0		& 0		& 0		& 0 	\\
		\textit{Hystrix cristata}			& 0		& 56	& 0		& 0		& 0		& 0		& 0		& 0		& 0		& 0		& 0		& 0		& 0		\\
		\textit{Crocuta crocuta}			& 0		& 0		& 42	& 0		& 1		& 0		& 0		& 0		& 0		& 0		& 0		& 0		& 0		\\
		\textit{Loxodonta africana}			& 0		& 0		& 0		& 149	& 0		& 0		& 0		& 3		& 0		& 0		& 0		& 3		& 0		\\
		\textit{Acinonyx jubatus}			& 0		& 0		& 0		& 0		& 22	& 0		& 0		& 0		& 0		& 0		& 0		& 0		& 0		\\
		\textit{Papio sp}					& 0		& 0		& 1		& 0		& 8		& 202	& 0		& 0		& 0		& 0		& 0		& 0		& 0		\\
		\textit{Blank}						& 0		& 0		& 0		& 0		& 0		& 0		& 289	& 0		& 0		& 0		& 0		& 0		& 0		\\
		\textit{Rhinocerotidae}				& 0		& 2		& 0		& 5		& 0		& 0		& 0		& 337	& 0		& 0		& 0		& 7		& 0		\\	
		\textit{Connochaetes taurinus}		& 0		& 0		& 4		& 17	& 0		& 9		& 0		& 3		& 615	& 9		& 0		& 2		& 3		\\
		\textit{Tragelaphus oryx}			& 2		& 0		& 3		& 1		& 0		& 0		& 0		& 8		& 11	& 316	& 2		& 7		& 1		\\
		\textit{Giraffa camelopardalis}		& 0		& 0		& 0		& 0		& 1		& 0		& 0		& 0		& 0		& 0		& 275	& 0		& 1		\\
		\textit{Panthera leo}				& 0		& 0		& 0		& 0		& 0		& 0		& 0		& 0		& 0		& 0		& 0		& 22	& 0		\\
		\textit{Equus quagga}				& 8		& 15	& 0		& 1		& 4		& 1		& 0		& 0		& 0		& 0		& 2		& 3		& 854	\\
		\bottomrule
	\end{tabularx}
\end{table}
\subsubsection{ROC and AUC for Sub-Saharan Classification Model}
The ROC in Figure 9 provides a visual assessment of the models inference results which indicates the model performed remarkably well for all animal classes as the AUC values are high. It can be concluded that the trained model in this trial achieved excellent results for each class. The plot and AUC values align with the outcomes presented in Table 5 and Table 6 which validates the deep learning aspects presented in this paper.
\begin{figure}[htp] 
	\centering
	\includegraphics[width=1\linewidth]{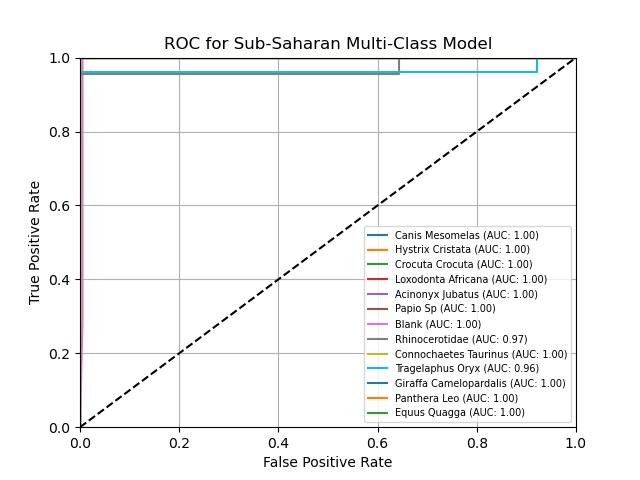}
	\caption{Receiver Operator Curve for Sub-Saharan Multi-Class Model}
	\label{fig9} 
\end{figure}
\subsubsection{BioPay}
The evaluation is concluded with the outcomes obtained from the BioPay service. Table 7 shows the ordered species detection counts collected during the 10-month trial. The \textit{Canis mesomelas} species had the lowest count with only 34 detections, whereas the \textit{Equus quagga} had the highest at 7158. During the trial, each detected animal initiated the transfer of 0.1 GBP from the respective species account to the guardian account. Following the trial's completion, the guardian earned £0.34 GBP for the \textit{Canis mesomelas}, £0.37 GBP for the \textit{Hystrix cristata}, £0.58 GBP for the \textit{Crocuta crocuta}, £1.48 GBP for the \textit{Loxodonta africana}, £2.22 GBP for the \textit{Acinonyx jubatus}, £7.48 GBP for the \textit{Papio sp}, £9.98 GBP for the \textit{Rhinocerotidae}, £10.22 GBP for the \textit{Connochaetes taurinus}, £10.58 GBP for the \textit{Tragelaphus oryx}, £26.46 GBP for the \textit{Giraffa camelopardalis}, £43.91 GBP for the \textit{Panthera leo}, and £71.58 GBP for the \textit{Equus quagga}. The guardians total earnings for the 10-month trial was £185.20 GBP.
\begin{table}[H] 
	\caption{Confusion Matrix for Sub-Saharan Model Inference.\label{tab5}}
	\newcolumntype{C}{>{\centering\arraybackslash}X}
	\begin{tabularx}{\textwidth}{p{3.5cm}CC}
		\toprule
		& \textbf{Detections}	& \textbf{Guardian Payment (GBP)} \\
		\midrule
		\textit{Canis mesomelas}		& 34		& 0.34		\\
		\textit{Hystrix cristata}		& 37		& 0.37		\\
		\textit{Crocuta crocuta}		& 58		& 0.58		\\
		\textit{Loxodonta africana}		& 148		& 1.48		\\
		\textit{Acinonyx jubatus}		& 222		& 2.22		\\
		\textit{Papio sp}				& 748		& 7.48		\\
		\textit{Rhinocerotidae}			& 998		& 9,98		\\	
		\textit{Connochaetes taurinus}	& 1022		& 10.22		\\
		\textit{Tragelaphus oryx}		& 1058		& 10.58		\\
		\textit{Giraffa camelopardalis}	& 2646		& 26.46		\\
		\textit{Panthera leo}			& 4391		& 43.91		\\
		\textit{Equus quagga}			& 7158		& 71.58		\\
		\bottomrule
		\textbf{Total}			& \textbf{18520}		& \textbf{185.20}    \\
		\bottomrule
	\end{tabularx}
\end{table} 

\section{Discussion}   
Using deep learning for species identification, and 3/4G camera traps for capturing images of animals, we successfully deployed a Sub-Saharan Africa model capable of detecting 12 distinct animal species. The deep learning model was trained with 1099 and 1771 tags per animal, which enabled us to effectively monitor a 400km${^2}$ region in Welgevonden Game Reserve in Limpopo Province in South Africa using 27 real-time 3/4G camera traps for a period of 10 months. During the model training phase it was possible to obtain good results for the RPN (objectness loss=0.0366, localisation loss=0.0112 for the training set and objectness loss=0.0530 and localisation loss=0.0384 for the validation set). The Box Classifier losses for classification and localisation were also good (classification loss=0.1833 and localisation loss=0.0261 for the training set and classification loss=0.1533 and localisation loss=0.0242 for the validation set). Combining the evaluation metrics, it was possible to obtain a total loss=0.2244 for the training set and a total loss=0.2690 for the validation set. Again, good results for an object detection model. The precision and recall results obtained for the validation set were also good with mAP=0.7542, mAP@.50IOU=0.9449 and mAP@.75IOU=0.8601, AR@1=0.6239, AR@100=0.8140 and AR@100(Large)=0.8357. The results show that the model can accurately identify objects in images with high precision and recall while maintaining a high level of localisation accuracy.  
\par
Throughout the 10 month trial, the Sensitivity (96.38\%), Specificity (99.6\%), Precision (87.14\%), F1 score (90.33\%), and Accuracy (99.31\%) metrics were consistently high for most species, thereby confirming the training results. However, the Precision scores for \textit{Hystrix cristata} and \textit{Acinonyx jubatus} were lower, with 77.09\% and 59.33\%, respectively. Figure 10, shows a sample image of a \textit{Hystrix cristata}, and indicates that they appear as small objects in the image, which aligns with the model's limited ability to detect small objects, as evidenced in Tables 3 and 4. Similarly, Figure 11, shows a sample image of \textit{Acinonyx jubatus}, which is also small, thereby making their detection more challenging, especially at night when the quality of images is lower. It is also worth noting that the model was trained using traditional camera trap data. Historically, conservationists have fixed camera traps much lower down (closer to the ground) so the animal appears larger in the image. As can be seen in figure 4 our camera trap deployments are much higher. This was done to prevent the cameras and solar panels being damaged by passing wildlife. Obviously, having cameras lower down removes issues where far away animals are misclassified as they would not be in the image. Camera deployment needs some further consideration.
\begin{figure}[htp] 
	\centering
	\includegraphics[width=0.88\linewidth]{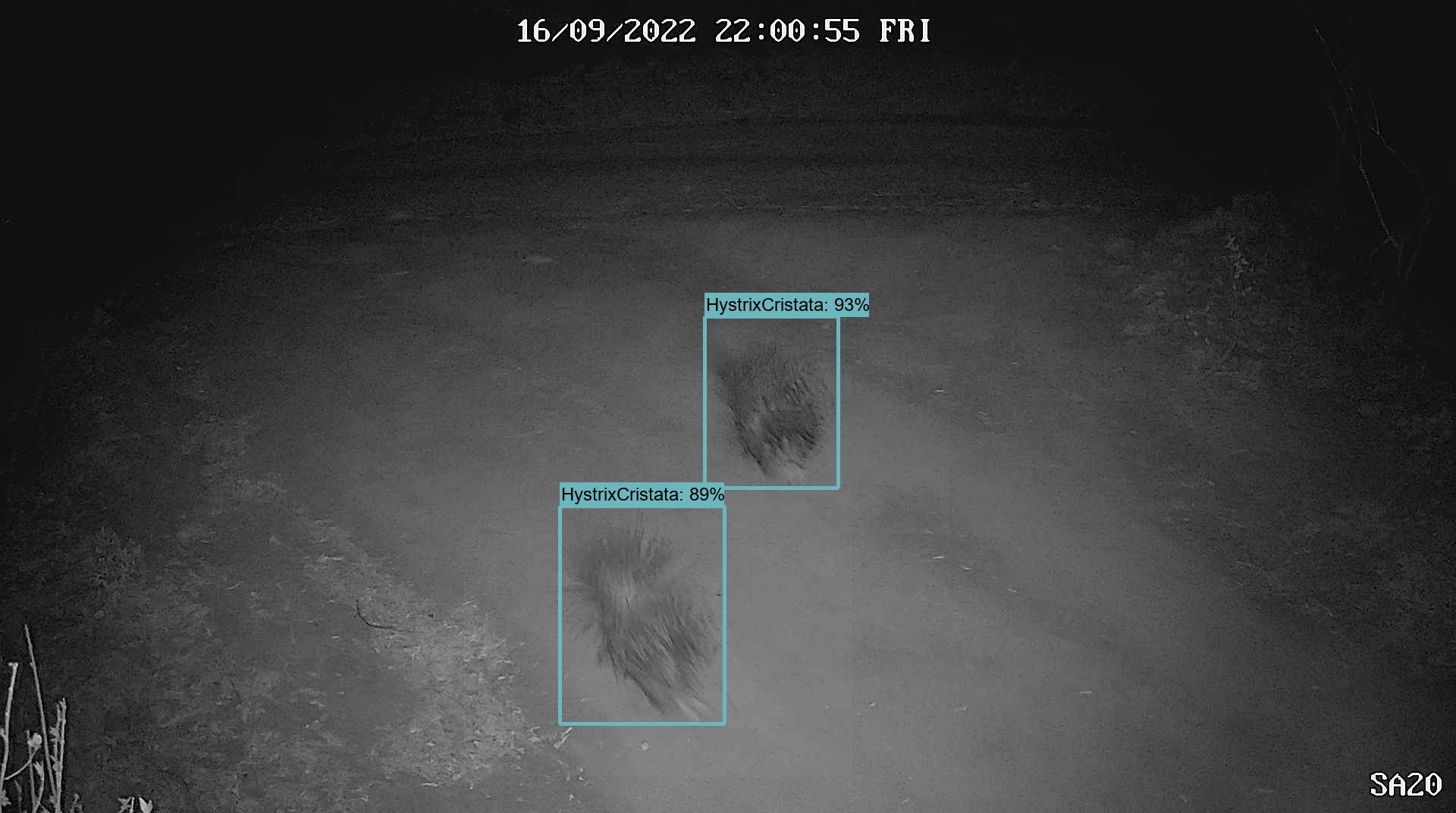}
	\caption{Shows two \textit{Hysterix cristata}: This is a particularly difficult detection given that the image was taken at night and the quality is poor. In this instance the model correctly detects these but it could just as easily classified tuffs of grass as a \textit{Hysterix cristata}}
	\label{fig10} 
\end{figure} 
\par
The \textit{Canis mesomelas}, \textit{Crocuta crocuta}, and \textit{Loxodonta africana} were also found to be misclassified as \textit{Acinonyx jubatus}, \textit{Rhinocerotidae}, and either \textit{Rhinocerotidae} or \textit{Panthera leo}, respectively, as evidenced in the confusion matrix in Table 6. \textit{Hystrix cristata} was never misclassified, but \textit{Equus quagga} and \textit{Rhinocerotidae} were mistakenly classified as \textit{Hystrix cristata}. \textit{Papio sp} was misclassified as either \textit{Crocuta crocuta} or \textit{Acinonyx jubatus}. Although \textit{Rhinocerotidae} performed well, it was also mistakenly classified as \textit{Loxodonta africana} or \textit{Panthera leo}, along with \textit{Hystrix cristata}. \textit{Tragelaphus oryx} and \textit{Connochaetes taurinus} were occasionally misclassified as each other. \textit{Giraffa camelopardalis} was identified correctly, and \textit{Panthera leo} was always correctly identified. \textit{Equus quagga} also performed well, but low-quality images captured at night sometimes resulted in them being misclassified as another species. To address this issue, the Sub-Saharan Africa model is continuously trained, and many of the incorrect classifications encountered during the trial have been identified and incorporated back into the training dataset to improve model performance. The version of the model used during the trial was version 18, the current version is 22.
\begin{figure}[htp] 
	\centering
	\includegraphics[width=0.88\linewidth]{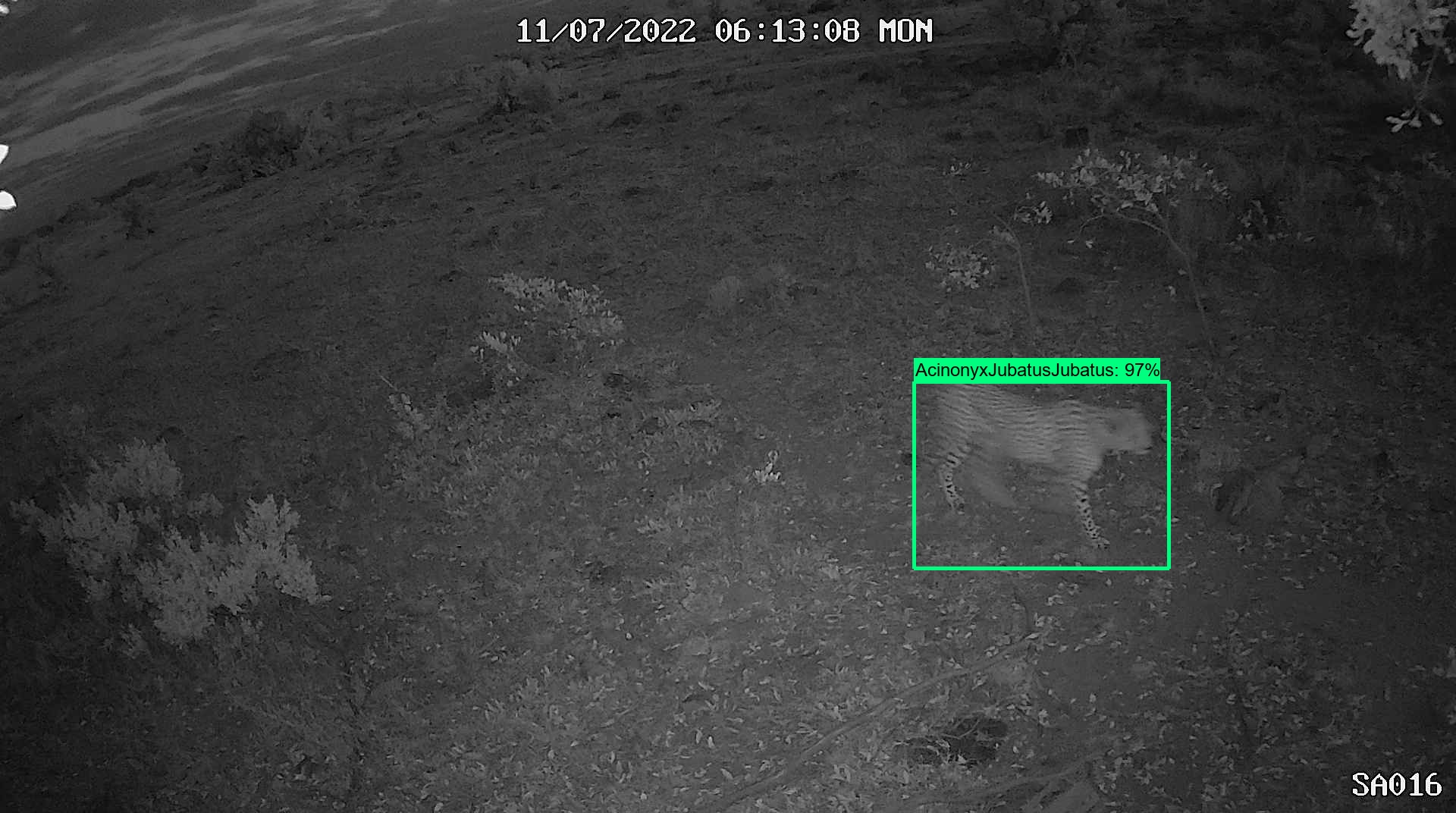}
	\caption{A rare image of a \textit{Acinonyx jubatus} in the trial which was the least captured animal over the 10-month period. Again, this is a very difficult visual of an animal that the model correctly detects}
	\label{fig11} 
\end{figure} 
\par 
Distance was also a problem in the study \cite{palencia2022towards}. Animals closest to the camera, as you would expect, classify much better than those farther away. In the trial, the camera trigger distance was set to "High," which allows objects to be detected up to 30 feet away (9 meters). Again, this configuration is not typical in traditional camera trap deployments. Obviously, detections farther way depend on the size of the animal. Detecting large animals, like a \textit{Loxodonta africana} or a \textit{Giraffa camelopardalis} are mostly successful, smaller animals such as a \textit{Papio sp} less so or at least the confidence scores are significantly lower. This can be seen in Figure 12 where the \textit{Panthera leo} has a lower confidence score than animals captured close up as shown in Figure 1 were \textit{Equus quaggas'} captured close up have a higher confidence score than those in the distance. Not having a distance protocol in the study impacted the inclusion criteria for evaluation. We evaluated farther detections that would not normally trigger the camera trap (animals closer to the camera were responsible for the trigger) and this likely impacted the results in some instances. Further investigation is needed to define a protocol to map distance and detection success and incorporate it into the ground-truth object selection criteria for evaluation. 
\par
The BioPay service performed as expected and the results show the successful transfer of funds between animals and the associated guardian. The detection results during inference show that overall detection success was high with a small number of miss-classifications. This would mean that money between species accounts would be made when in fact that animal was not actually seen. This will always be a difficult challenge to address, but a small margin of error in this case is negligible. Obviously, this may not be the case when species are appropriately valued where highly prized animals could transfer large amounts of money when they are misclassified. This will need to be considered in future studies. 
\par
Another important point to raise is the incentive surrounding the monetary gain guardians would get for caring for animals compared to the amount obtained for poaching. Obviously, the former would have to be much higher if something like BioPay is to be given a chance of success. In other words receiving £100 pounds a month to ensure the safety of a \textit{Rhinocerotidae} compared to a few thousand pounds they would receive for its horn (poachers lower down the IWT chain get significantly less than those close to the source of the sale \cite{duffy2013poverty}) would unlikely be attractive to those involved in poaching. Another factor to be considered is land opportunity costs \cite{ayompe2021does}. If alternatives to conservation are very profitable (e.g., oil palm), then payment for species' presence would also need to be much higher to be effective. 
\begin{figure}[htp] 
	\centering
	\includegraphics[width=0.88\linewidth]{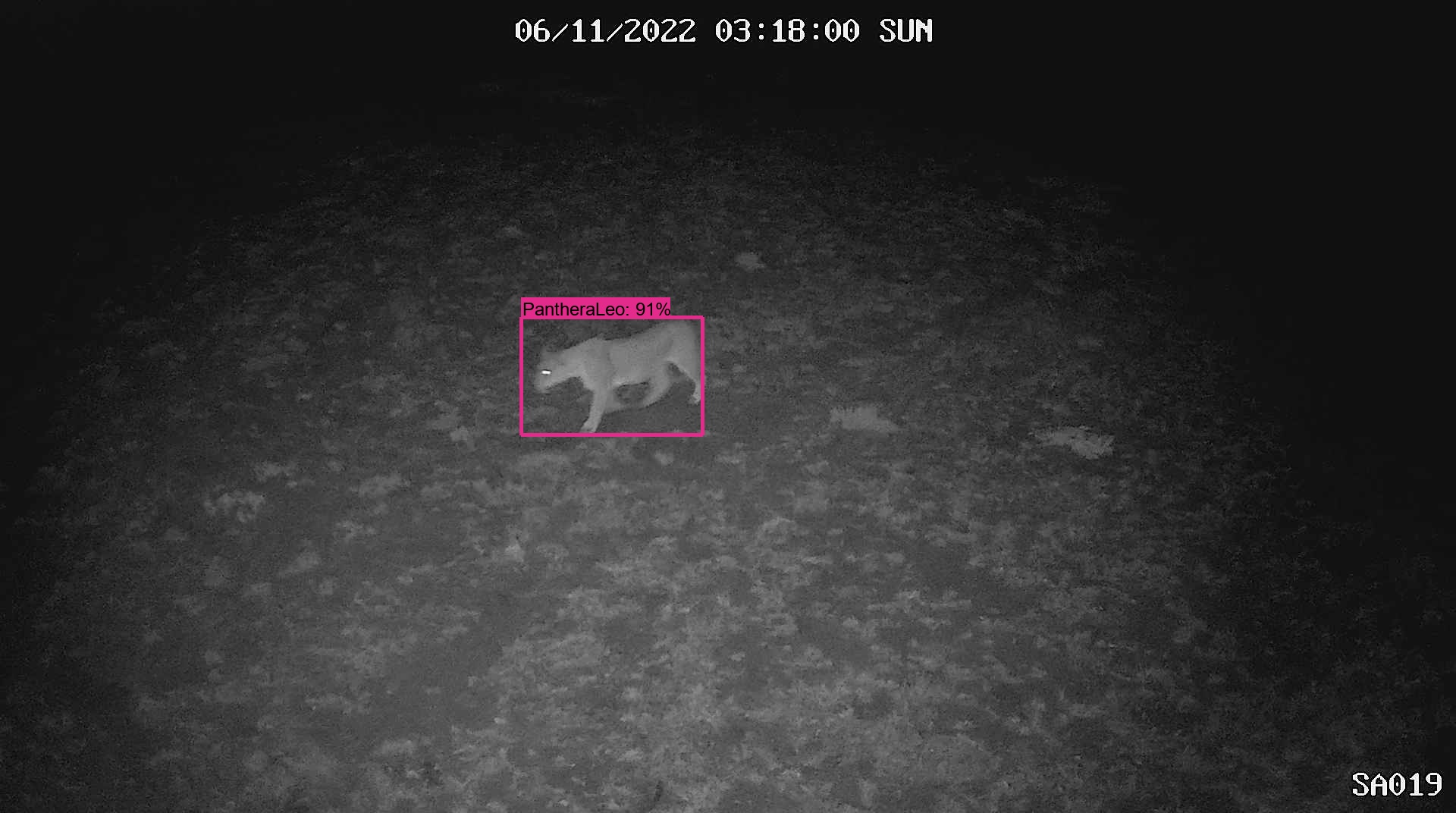}
	\caption{A rare image of a \textit{Panthera leo} in the trial which is also a very difficult detection that the model correctly detects}
	\label{fig9} 
\end{figure}
\par
Despite these issues, the results were encouraging. To the best of our knowledge this is the first extensive evaluation we know of that combines deep learning, 3/4G camera traps to monitor animal populations in real-time and provision a monetary reward scheme for guardians. We acknowledge that the trial was limited in scope and that we would need to significantly increase the number of camera traps used in the study as well as increase the number species in the Sub-Saharan Africa model, for example, \textit{Potamochoerus larvatus} and \textit{Panthera pardus}, which were captured during the trial. However, poor 3/4G signal and the ongoing destruction of camera traps by \textit{Loxodonta africana} will likely affect scale up. Cases to protect camera traps from animals, may help, and in other sites they will need to be protected from being stolen (by humans). 
\par
We also recognise that a much longer study period is needed to fully evaluate the approach and connect BioPay to real monetary systems. However, an independent study would have to be conducted to value species before BioPay could be fully implemented \cite{costanza2020valuing}. Some of the factors to consider would be: 1) conservation status and rarity especially if endangered and threatened \cite{talukdar2018conservation}; 2) economic value in terms of ecotourism potential and medicinal value could influence their perceived value \cite{courchamp2006rarity}; 3) ecological value such as their role in maintaining ecosystem health or providing ecosystem services. For example, white rhinoceros, in addition to being of high tourism value they are also facilitators, providing other grazing herbivores, with improved grazing conditions \cite{waldram2008ecological}; 4) cultural significance can impact perceived value, for example, by being considered sacred or having an important role in traditional cultural practices \cite{berkes2017sacred}; and 5) local knowledge about mammal behaviour, ecology and uses could influence their perceived value \cite{pierotti2010indigenous}. 
\par
Finally, any credible study would need to include guardians and there would need to be associated stewardship and biodiversity measurement protocols to fully evaluate the utility and impact of the system - this will required some serious thinking. Nonetheless, we believe that this work, inspired by "Interspecies Money", provides a working blueprint and the necessary evidence to support a much bigger study.

\section{Conclusions}

This paper introduced an equitable digital stewardship and reward system for wildlife guardians that utilises deep learning and 3/4G camera traps to detect animal species in real-time. This provides a blueprint that allows local stakeholders to be rewarded for the welfare services they provide. The findings are encouraging and show that distinct species in images can be detected with high accuracy \cite{russakovsky2015imagenet}. Several similar species detection studies have been reported in the literature \cite{swanson2015snapshot}, \cite{tabak2019machine}, \cite{willi2019identifying}, \cite{yousif2017fast}, \cite{norouzzadeh2018automatically}, \cite{norouzzadeh2021deep}, \cite{villa2017towards}. However, the central focus differs to that presented in this paper. By combining deep learning and 3/4G camera traps, analysis occurs as a single unified process that records and raises alerts. This allows services to be superimposed on top of this technology to derive insights in real-time and promote new innovations in conservation.
\par
We proposed a BioPay service in this paper that builds on this idea. It is a disruptive but necessary service that aims to include and reward local stakeholders for the stewardship services they provide. As the literature in this paper indicates, local guardians are seen as crucial components in successful conservation \cite{reyes2022recognizing}. However, the barrier to their success has been that they continue to face challenges with full participation in the crafting and implementation of biodiversity policy at local, regional, and global levels and as such are poorly compensated for the services they provide \cite{witter2015moments}. Many initiatives have excluded local stakeholders through management regimes that outlaw local practices and customary institutions \cite{dawson2021role}. Yet, the findings have shown that attempts to separate biodiversity and local livelihoods have yielded limited success: biodiversity often declines at the same time as the well-being of those who inhabit areas targeted for interventions \cite{sachedina2010disconnected}. BioPay includes and rewards local people, particular the poorest among them, for the services they provide as and when animal species are detected within regions requiring biodiversity support. 
\par
The solution is rudimentary and we acknowledge that implementing BioPay at scale will be difficult as it is not clear who actually receives payment. For example, should it go to the community or only people who take an active interest in the care and provision of services for animals that live in the locale. It may be best to let communities themselves decide who should be rewarded and how funds are spent \cite{ledgard2022interspecies}. Whatever the approach, payments will be conditional on the continued presence of species. Payments to guardians responsible for monitoring presence would make such a system scalable. Guardians could redistribute the funds to those people that are in a position to ensure that species or its habitat prevail. BioPay will not fully address the complex nature of conservation and biodiversity management, but it may provide a tool to help redress the disproportionate allocation of global conservation funding by providing an equitable revenue sharing scheme that includes and rewards local stakeholders for the services they provide. 
\par
Having a service like BioPay may help to forge much closer relationship between guardians of animal welfare, governments and NGOs and improve conservation outcomes. An alternative view however might be to bypass governments and NGOs altogether and use automated (blockchain \cite{zheng2018blockchain}) to directly pay guardians, which would make their roles less relevant and less in need of financing. This may certainly increase the 1\% allocation towards nature-based solutions we refer to earlier in the paper. COP26 recognised the need to reward local stakeholders. However, as we pointed out in this paper, the US\$1.7 billion allocated is a fraction of the US\$124-143 billion allocated annually to organisations working in conservation. These funds rarely reach the poorest in local communities who are in most need of support \cite{dowie2011conservation}. We believe that the findings in this paper provide a viable blueprint based on the "Interspecies Money" principal that will facilitate the transfer of funds between animals and their associated guardian groups.   
\par
Despite the encouraging results however there is a great deal of future work needed. The size of the study was insufficient to fully understand the complete set of requirements needed to implement the BioPay revenue sharing scheme. A much larger representation of animals in the model is required to ensure equal representation for all animals in the environment being monitored. There also needs to be a detailed assessment to understand what each species is worth. There were no recipients for the BioPay funds transferred in the study and a model for including local stakeholders would need to be defined as well as a clear understand of who gets what money and under what circumstances this happens. 
\par
Conservation AI is a growing platform which already has 28 active studies worldwide. At the time of writing it has processed more than 5 million images, in just over 12 months, from 75 real-time cameras and historical datasets uploaded by partners. In the next 12 months we anticipate significant growth and this will allow us to run increasingly bigger studies to help us to address the limitations highlighted in this paper.
\par
Overall, the results show the potential which we think warrants further investigation. This work is multidisciplinary and contributes to the machine learning and conservation fields. We hope that the study provides new insights on how deep learning algorithms combined with 3/4G camera traps can be used to measure and monitor biodiversity health and provide revenue sharing schemes that benefit guardians for the wildlife and biodiversity services they provide. 

\section{Acknowledgements} 
The authors would like to thank Welgevonden Game Reserve in Limpopo Province in South Africa for allowing us to visit and install the camera traps for the trial. We would like to thank Reolink for donating the camera traps to us and Vodafone in the UK and Vodacom in South Africa for sponsoring our communications. We would like to thank Knowsley Safari in Merseyside in the UK for allowing us to install cameras to collect the data needed to train the Sub-Saharan model to detect the animals monitored in the study. Finally, the authors would like to thank Mrs Rachel Chalmers for the significant amount of work she has done over the last four years for tagging the species in Conservation AI.  


\begin{adjustwidth}{-\extralength}{0cm}

\reftitle{References}


\bibliography{litrev1}

\begin{thebibliography}{999}

\bibitem[Mora \em{et~al.}(2011)Mora, Tittensor, Adl, Simpson, and
  Worm]{mora2011many}
Mora, C.; Tittensor, D.P.; Adl, S.; Simpson, A.G.; Worm, B.
\newblock How many species are there on Earth and in the ocean?
\newblock {\em PLoS biology} {\bf 2011}, {\em 9},~e1001127.

\bibitem[Nations(2019)]{united2019report}
Nations, U.
\newblock UN Report: Nature’s Dangerous Decline ‘Unprecedented’; Species
  Extinction Rates ‘Accelerating’.
\newblock {\em Sustainable Development Goals} {\bf 2019}.

\bibitem[Andermann \em{et~al.}(2020)Andermann, Faurby, Turvey, Antonelli, and
  Silvestro]{andermann2020past}
Andermann, T.; Faurby, S.; Turvey, S.T.; Antonelli, A.; Silvestro, D.
\newblock The past and future human impact on mammalian diversity.
\newblock {\em Science Advances} {\bf 2020}, {\em 6},~eabb2313.

\bibitem[Pereira \em{et~al.}(2012)Pereira, Navarro, and
  Martins]{pereira2012global}
Pereira, H.M.; Navarro, L.M.; Martins, I.S.
\newblock Global biodiversity change: the bad, the good, and the unknown.
\newblock {\em Annual Review of Environment and Resources} {\bf 2012}, {\em
  37},~25--50.

\bibitem[Ellis(2013)]{ellis2013tiger}
Ellis, R.
\newblock {\em Tiger bone \& rhino horn: the destruction of wildlife for
  traditional Chinese medicine}; Island Press,  2013.

\bibitem[Weru(2016)]{weru2016wildlife}
Weru, S.
\newblock Wildlife protection and trafficking assessment in Kenya: Drivers and
  trends of transnational wildlife crime in Kenya and its role as a transit
  point for trafficked species in East Africa (PDF, 3.5 MB {\bf 2016}.

\bibitem[(UNEP) and INTERPOL(2016)]{united2016report}
(UNEP), U.N.E.P.; INTERPOL.
\newblock UNEP-INTERPOL report: value of environmental crime up 26
\newblock {\em Environmental Rights AND Governance} {\bf 2016}.

\bibitem[Gonzalez~Estrada(2022)]{gonzalez2022influence}
Gonzalez~Estrada, A.J.
\newblock The influence of illicit wildlife trafficking in security matters.
  The case of illicit trafficking of elephant ivory and rhino horn in Africa.
\newblock Master's thesis, UiT Norges arktiske universitet,  2022.

\bibitem[McClenachan \em{et~al.}(2016)McClenachan, Cooper, and
  Dulvy]{mcclenachan2016rethinking}
McClenachan, L.; Cooper, A.B.; Dulvy, N.K.
\newblock Rethinking trade-driven extinction risk in marine and terrestrial
  megafauna.
\newblock {\em Current Biology} {\bf 2016}, {\em 26},~1640--1646.

\bibitem[Eikelboom \em{et~al.}(2020)Eikelboom, Nuijten, Wang, Schroder,
  Heitk{\"o}nig, Mooij, van Langevelde, and Prins]{eikelboom2020will}
Eikelboom, J.A.; Nuijten, R.J.; Wang, Y.X.; Schroder, B.; Heitk{\"o}nig, I.M.;
  Mooij, W.M.; van Langevelde, F.; Prins, H.H.
\newblock Will legal international rhino horn trade save wild rhino
  populations?
\newblock {\em Global ecology and conservation} {\bf 2020}, {\em 23},~e01145.

\bibitem[Sharma \em{et~al.}(2020)Sharma, Sharma, Katuwal, Chaulagain, and
  Belant]{sharma2020people}
Sharma, S.; Sharma, H.P.; Katuwal, H.B.; Chaulagain, C.; Belant, J.L.
\newblock People’s knowledge of illegal Chinese pangolin trade routes in
  central Nepal.
\newblock {\em Sustainability} {\bf 2020}, {\em 12},~4900.

\bibitem[McKirdy(2019)]{mckirdy2019record}
McKirdy, E.
\newblock Record Haul of Pangolin Scales Highlights Chinese and Vietnamese
  Demand for Endangered Species.
\newblock {\em CNN News. April} {\bf 2019}, {\em 12},~2019.

\bibitem[Raustiala(1997)]{raustiala1997states}
Raustiala, K.
\newblock States, NGOs, and international environmental institutions.
\newblock {\em International Studies Quarterly} {\bf 1997}, {\em 41},~719--740.

\bibitem[White \em{et~al.}(2022)White, Petrovan, Christie, Martin, and
  Sutherland]{white2022price}
White, T.B.; Petrovan, S.O.; Christie, A.P.; Martin, P.A.; Sutherland, W.J.
\newblock What is the Price of Conservation? A Review of the Status Quo and
  Recommendations for Improving Cost Reporting.
\newblock {\em BioScience} {\bf 2022}.

\bibitem[Girardin \em{et~al.}(2021)Girardin, Jenkins, Seddon, Allen, Lewis,
  Wheeler, Griscom, and Malhi]{girardin2021nature}
Girardin, C.A.; Jenkins, S.; Seddon, N.; Allen, M.; Lewis, S.L.; Wheeler, C.E.;
  Griscom, B.W.; Malhi, Y.
\newblock Nature-based solutions can help cool the planet—if we act now.
\newblock {\em Nature} {\bf 2021}, {\em 593},~191--194.

\bibitem[Holmes(2012)]{holmes2012biodiversity}
Holmes, G.
\newblock Biodiversity for billionaires: capitalism, conservation and the role
  of philanthropy in saving/selling nature.
\newblock {\em Development and change} {\bf 2012}, {\em 43},~185--203.

\bibitem[Wang and Zhi(2016)]{wang2016role}
Wang, Y.; Zhi, Q.
\newblock The role of green finance in environmental protection: Two aspects of
  market mechanism and policies.
\newblock {\em Energy Procedia} {\bf 2016}, {\em 104},~311--316.

\bibitem[Linton \em{et~al.}(2007)Linton, Klassen, and
  Jayaraman]{linton2007sustainable}
Linton, J.D.; Klassen, R.; Jayaraman, V.
\newblock Sustainable supply chains: An introduction.
\newblock {\em Journal of operations management} {\bf 2007}, {\em
  25},~1075--1082.

\bibitem[Blunt \em{et~al.}(2011)Blunt, Turner, and Hertz]{blunt2011meaning}
Blunt, P.; Turner, M.; Hertz, J.
\newblock The meaning of development assistance.
\newblock {\em Public Administration and Development} {\bf 2011}, {\em
  31},~172--187.

\bibitem[Bull \em{et~al.}(2013)Bull, Suttle, Gordon, Singh, and
  Milner-Gulland]{bull2013biodiversity}
Bull, J.W.; Suttle, K.B.; Gordon, A.; Singh, N.J.; Milner-Gulland, E.
\newblock Biodiversity offsets in theory and practice.
\newblock {\em Oryx} {\bf 2013}, {\em 47},~369--380.

\bibitem[da~Silva and Wheeler(2017)]{da2017ecosystems}
da~Silva, J.M.C.; Wheeler, E.
\newblock Ecosystems as infrastructure.
\newblock {\em Perspectives in ecology and conservation} {\bf 2017}, {\em
  15},~32--35.

\bibitem[Pretty \em{et~al.}(2001)Pretty, Brett, Gee, Hine, Mason, Morison,
  Rayment, Van Der~Bijl, and Dobbs]{pretty2001policy}
Pretty, J.; Brett, C.; Gee, D.; Hine, R.; Mason, C.; Morison, J.; Rayment, M.;
  Van Der~Bijl, G.; Dobbs, T.
\newblock Policy challenges and priorities for internalizing the externalities
  of modern agriculture.
\newblock {\em Journal of environmental planning and management} {\bf 2001},
  {\em 44},~263--283.

\bibitem[Estrada \em{et~al.}(2022)Estrada, Garber, Gouveia,
  Fern{\'a}ndez-Llamazares, Ascens{\~a}o, Fuentes, Garnett, Shaffer,
  Bicca-Marques, Fa, et~al.]{estrada2022global}
Estrada, A.; Garber, P.A.; Gouveia, S.; Fern{\'a}ndez-Llamazares, {\'A}.;
  Ascens{\~a}o, F.; Fuentes, A.; Garnett, S.T.; Shaffer, C.; Bicca-Marques, J.;
  Fa, J.E.;  et~al.
\newblock Global importance of Indigenous Peoples, their lands, and knowledge
  systems for saving the world’s primates from extinction.
\newblock {\em Science advances} {\bf 2022}, {\em 8},~eabn2927.

\bibitem[Turner \em{et~al.}(2012)Turner, Brandon, Brooks, Gascon, Gibbs,
  Lawrence, Mittermeier, and Selig]{turner2012global}
Turner, W.R.; Brandon, K.; Brooks, T.M.; Gascon, C.; Gibbs, H.K.; Lawrence,
  K.S.; Mittermeier, R.A.; Selig, E.R.
\newblock Global biodiversity conservation and the alleviation of poverty.
\newblock {\em BioScience} {\bf 2012}, {\em 62},~85--92.

\bibitem[Ledgard(2022)]{ledgard2022interspecies}
Ledgard, J.
\newblock Interspecies Money.
\newblock {\em Breakthrough: The Promise of Frontier Technologies for
  Sustainable Development} {\bf 2022}, p.~77.

\bibitem[Reyes-Garc{\'\i}a \em{et~al.}(2022)Reyes-Garc{\'\i}a,
  Fern{\'a}ndez-Llamazares, Aumeeruddy-Thomas, Benyei, Bussmann, Diamond,
  Garc{\'\i}a-Del-Amo, Guadilla-S{\'a}ez, Hanazaki, Kosoy,
  et~al.]{reyes2022recognizing}
Reyes-Garc{\'\i}a, V.; Fern{\'a}ndez-Llamazares, {\'A}.; Aumeeruddy-Thomas, Y.;
  Benyei, P.; Bussmann, R.W.; Diamond, S.K.; Garc{\'\i}a-Del-Amo, D.;
  Guadilla-S{\'a}ez, S.; Hanazaki, N.; Kosoy, N.;  et~al.
\newblock Recognizing Indigenous peoples’ and local communities’ rights and
  agency in the post-2020 Biodiversity Agenda.
\newblock {\em Ambio} {\bf 2022}, {\em 51},~84--92.

\bibitem[Dawson \em{et~al.}(2021)Dawson, Coolsaet, Sterling, Loveridge, Nicole,
  Wongbusarakum, Sangha, Scherl, Phan, Zafra-Calvo, et~al.]{dawson2021role}
Dawson, N.; Coolsaet, B.; Sterling, E.; Loveridge, R.; Nicole, D.;
  Wongbusarakum, S.; Sangha, K.; Scherl, L.; Phan, H.P.; Zafra-Calvo, N.;
  et~al.
\newblock The role of Indigenous peoples and local communities in effective and
  equitable conservation.
\newblock {\em Ecology and Society} {\bf 2021}, {\em 26}.

\bibitem[Ruckelshaus \em{et~al.}(2020)Ruckelshaus, Jackson, Mooney, Jacobs,
  Kassam, Arroyo, B{\'a}ldi, Bartuska, Boyd, Joppa,
  et~al.]{ruckelshaus2020ipbes}
Ruckelshaus, M.H.; Jackson, S.T.; Mooney, H.A.; Jacobs, K.L.; Kassam, K.A.S.;
  Arroyo, M.T.; B{\'a}ldi, A.; Bartuska, A.M.; Boyd, J.; Joppa, L.N.;  et~al.
\newblock The IPBES global assessment: pathways to action.
\newblock {\em Trends in Ecology \& Evolution} {\bf 2020}, {\em 35},~407--414.

\bibitem[Haenssgen \em{et~al.}(2022)Haenssgen, Lechner, Rakotonarivo,
  Leepreecha, Sakboon, Chu, Auclair, and Vlaev]{haenssgen2022implementation}
Haenssgen, M.J.; Lechner, A.M.; Rakotonarivo, S.; Leepreecha, P.; Sakboon, M.;
  Chu, T.W.; Auclair, E.; Vlaev, I.
\newblock Implementation of the COP26 declaration to halt forest loss must
  safeguard and include Indigenous people.
\newblock {\em Nature Ecology \& Evolution} {\bf 2022}, {\em 6},~235--236.

\bibitem[Bandiaky-Badji \em{et~al.}(2023)Bandiaky-Badji, Lovera, M{\'a}rquez,
  Leiva, Robinson, Smith, Currey, Ross, Agrawal, and
  White]{bandiaky2023indigenous}
Bandiaky-Badji, S.; Lovera, S.; M{\'a}rquez, G.Y.H.; Leiva, F.J.A.; Robinson,
  C.J.; Smith, M.A.; Currey, K.; Ross, H.; Agrawal, A.; White, A.
\newblock Indigenous stewardship for habitat protection.
\newblock {\em One Earth} {\bf 2023}, {\em 6},~68--72.

\bibitem[Laird and Wynberg()]{lairdconnecting}
Laird, S.; Wynberg, R.
\newblock CONNECTING THE DOTS... BIODIVERSITY CONSERVATION, SUSTAINABLE USE.

\bibitem[Sharef \em{et~al.}(2022)Sharef, Nasharuddin, Mohamed, Zamani, Osman,
  and Yaakob]{sharef2022applications}
Sharef, N.M.; Nasharuddin, N.A.; Mohamed, R.; Zamani, N.W.; Osman, M.H.;
  Yaakob, R.
\newblock Applications of Data Analytics and Machine Learning for Digital
  Twin-based Precision Biodiversity: A Review.
\newblock In Proceedings of the 2022 International Conference on Advanced
  Creative Networks and Intelligent Systems (ICACNIS). IEEE,  2022, pp. 1--7.

\bibitem[Escobar(1998)]{escobar1998whose}
Escobar, A.
\newblock Whose knowledge, whose nature? Biodiversity, conservation, and the
  political ecology of social movements.
\newblock {\em Journal of political ecology} {\bf 1998}, {\em 5},~53--82.

\bibitem[Chesson(2000)]{chesson2000mechanisms}
Chesson, P.
\newblock Mechanisms of maintenance of species diversity.
\newblock {\em Annual review of Ecology and Systematics} {\bf 2000}, pp.
  343--366.

\bibitem[Parry(2010)]{parry2010age}
Parry, J.H.
\newblock {\em The Age of Reconnaissance: Discovery, Exporation and Settlement,
  1450-1650}; Hachette UK,  2010.

\bibitem[Dunlap(1988)]{10.2307/3984377}
Dunlap, T.R.
\newblock Sport Hunting and Conservation, 1880-1920.
\newblock {\em Environmental Review: ER} {\bf 1988}, {\em 12},~51--60.

\bibitem[Shaw(2021)]{shaw2021indigenous}
Shaw, C.
\newblock Indigenous and Community Conserved Areas.
\newblock {\em Environmental Defenders: Deadly Struggles for Life and
  Territory} {\bf 2021}, p.~80.

\bibitem[Hernandez(2022)]{hernandez2022fresh}
Hernandez, J.
\newblock {\em Fresh banana leaves: healing indigenous landscapes through
  indigenous science}; North Atlantic Books,  2022.

\bibitem[Runte(1997)]{runte1997national}
Runte, A.
\newblock {\em National parks: the American experience}; U of Nebraska Press,
  1997.

\bibitem[Oguamanam(2022)]{oguamanam2022indigenous}
Oguamanam, C.
\newblock Indigenous peoples rights in equitable benefit-sharing over genetic
  resources: digital sequence information (DSI) and a new technological
  landscape. In {\em Research Handbook on the International Law of Indigenous
  Rights}; Edward Elgar Publishing,  2022; pp. 354--375.

\bibitem[Cornell and Kalt(1992)]{cornell1992can}
Cornell, S.E.; Kalt, J.P.
\newblock {\em What can tribes do?: Strategies and institutions in American
  Indian economic development}; American Indian Studies Center, University of
  California, Los Angeles Los~…,  1992.

\bibitem[Dom{\'\i}nguez and Luoma(2020)]{dominguez2020decolonising}
Dom{\'\i}nguez, L.; Luoma, C.
\newblock Decolonising conservation policy: How colonial land and conservation
  ideologies persist and perpetuate indigenous injustices at the expense of the
  environment.
\newblock {\em Land} {\bf 2020}, {\em 9},~65.

\bibitem[Cooney \em{et~al.}(2018)Cooney, Roe, Dublin, Booker,
  et~al.]{cooney2018wild}
Cooney, R.; Roe, D.; Dublin, H.; Booker, F.;  et~al.
\newblock Wild Life, Wild Livelihoods: Involving communities on Sustainable
  Wildlife Management and Combating illegal Wildlife Trade {\bf 2018}.

\bibitem[Cooney and Challender(2020)]{cooney2020engaging}
Cooney, R.; Challender, D.W.
\newblock Engaging local communities in responses to illegal trade in
  pangolins: who, why and how? In {\em Pangolins}; Elsevier,  2020; pp.
  369--383.

\bibitem[Lyver \em{et~al.}(2019)Lyver, Timoti, Davis, and
  Tylianakis]{lyver2019biocultural}
Lyver, P.; Timoti, P.; Davis, T.; Tylianakis, J.
\newblock Biocultural hysteresis inhibits adaptation to environmental change.
\newblock {\em Trends in Ecology \& Evolution} {\bf 2019}, {\em 34},~771--780.

\bibitem[Ashton \em{et~al.}(1997)Ashton et~al.]{ashton1997industrial}
Ashton, T.S.;  et~al.
\newblock The industrial revolution 1760-1830.
\newblock {\em OUP Catalogue} {\bf 1997}.

\bibitem[Hawken \em{et~al.}(2013)Hawken, Lovins, and Lovins]{hawken2013natural}
Hawken, P.; Lovins, A.B.; Lovins, L.H.
\newblock {\em Natural capitalism: The next industrial revolution}; Routledge,
  2013.

\bibitem[Roser \em{et~al.}(2013)Roser, Ritchie, and
  Ortiz-Ospina]{roser2013world}
Roser, M.; Ritchie, H.; Ortiz-Ospina, E.
\newblock World population growth.
\newblock {\em Our world in data} {\bf 2013}.

\bibitem[Schmink and Wood(2019)]{schmink2019political}
Schmink, M.; Wood, C.H.
\newblock The “political ecology” of Amazonia. In {\em Lands at risk in the
  Third World: Local-level perspectives}; Routledge,  2019; pp. 38--57.

\bibitem[Tilman \em{et~al.}(1994)Tilman, May, Lehman, and
  Nowak]{tilman1994habitat}
Tilman, D.; May, R.M.; Lehman, C.L.; Nowak, M.A.
\newblock Habitat destruction and the extinction debt.
\newblock {\em Nature} {\bf 1994}, {\em 371},~65--66.

\bibitem[Nogu{\'e}s-Bravo \em{et~al.}(2008)Nogu{\'e}s-Bravo, Rodr{\'\i}guez,
  Hortal, Batra, and Ara{\'u}jo]{nogues2008climate}
Nogu{\'e}s-Bravo, D.; Rodr{\'\i}guez, J.; Hortal, J.; Batra, P.; Ara{\'u}jo,
  M.B.
\newblock Climate change, humans, and the extinction of the woolly mammoth.
\newblock {\em PLoS biology} {\bf 2008}, {\em 6},~e79.

\bibitem[Martin(2005)]{martin2005twilight}
Martin, P.S.
\newblock {\em Twilight of the mammoths: Ice Age extinctions and the rewilding
  of America}; Vol.~8, Univ of California Press,  2005.

\bibitem[Heintzman \em{et~al.}(2015)Heintzman, Zazula, Cahill, Reyes, MacPhee,
  and Shapiro]{heintzman2015genomic}
Heintzman, P.D.; Zazula, G.D.; Cahill, J.A.; Reyes, A.V.; MacPhee, R.D.;
  Shapiro, B.
\newblock Genomic data from extinct North American Camelops revise camel
  evolutionary history.
\newblock {\em Molecular Biology and Evolution} {\bf 2015}, {\em
  32},~2433--2440.

\bibitem[Diamond(1989)]{diamond1989present}
Diamond, J.M.
\newblock The present, past and future of human-caused extinctions.
\newblock {\em Philosophical Transactions of the Royal Society of London. B,
  Biological Sciences} {\bf 1989}, {\em 325},~469--477.

\bibitem[Anderson(1989)]{anderson1989mechanics}
Anderson, A.
\newblock Mechanics of overkill in the extinction of New Zealand moas.
\newblock {\em Journal of Archaeological Science} {\bf 1989}, {\em
  16},~137--151.

\bibitem[Perez \em{et~al.}(2005)Perez, Godfrey, Nowak-Kemp, Burney,
  Ratsimbazafy, and Vasey]{perez2005evidence}
Perez, V.R.; Godfrey, L.R.; Nowak-Kemp, M.; Burney, D.A.; Ratsimbazafy, J.;
  Vasey, N.
\newblock Evidence of early butchery of giant lemurs in Madagascar.
\newblock {\em Journal of Human Evolution} {\bf 2005}, {\em 49},~722--742.

\bibitem[Ceballos \em{et~al.}(2010)Ceballos, Garc{\'\i}a, and
  Ehrlich]{ceballos2010sixth}
Ceballos, G.; Garc{\'\i}a, A.; Ehrlich, P.R.
\newblock The sixth extinction crisis: Loss of animal populations and species.
\newblock {\em Journal of Cosmology} {\bf 2010}, {\em 8},~31.

\bibitem[Cowie \em{et~al.}(2022)Cowie, Bouchet, and Fontaine]{cowie2022sixth}
Cowie, R.H.; Bouchet, P.; Fontaine, B.
\newblock The Sixth Mass Extinction: fact, fiction or speculation?
\newblock {\em Biological Reviews} {\bf 2022}, {\em 97},~640--663.

\bibitem[Barnosky \em{et~al.}(2011)Barnosky, Matzke, Tomiya, Wogan, Swartz,
  Quental, Marshall, McGuire, Lindsey, Maguire, et~al.]{barnosky2011has}
Barnosky, A.D.; Matzke, N.; Tomiya, S.; Wogan, G.O.; Swartz, B.; Quental, T.B.;
  Marshall, C.; McGuire, J.L.; Lindsey, E.L.; Maguire, K.C.;  et~al.
\newblock Has the Earth’s sixth mass extinction already arrived?
\newblock {\em Nature} {\bf 2011}, {\em 471},~51--57.

\bibitem[Wiedmann \em{et~al.}(2020)Wiedmann, Lenzen, Key{\ss}er, and
  Steinberger]{wiedmann2020scientists}
Wiedmann, T.; Lenzen, M.; Key{\ss}er, L.T.; Steinberger, J.K.
\newblock Scientists’ warning on affluence.
\newblock {\em Nature communications} {\bf 2020}, {\em 11},~1--10.

\bibitem[Brown and Cameron(2000)]{brown2000can}
Brown, P.M.; Cameron, L.D.
\newblock What can be done to reduce overconsumption?
\newblock {\em Ecological Economics} {\bf 2000}, {\em 32},~27--41.

\bibitem[Opoku(2019)]{opoku2019biodiversity}
Opoku, A.
\newblock Biodiversity and the built environment: Implications for the
  Sustainable Development Goals (SDGs).
\newblock {\em Resources, conservation and recycling} {\bf 2019}, {\em
  141},~1--7.

\bibitem[Almond \em{et~al.}(2020)Almond, Grooten, and
  Peterson]{almond2020living}
Almond, R.E.; Grooten, M.; Peterson, T.
\newblock {\em Living Planet Report 2020-Bending the curve of biodiversity
  loss}; World Wildlife Fund,  2020.

\bibitem[Welford(2013)]{welford2013hijacking}
Welford, R.
\newblock {\em Hijacking environmentalism: Corporate responses to sustainable
  development}; Routledge,  2013.

\bibitem[Helm(2015)]{helm2015natural}
Helm, D.
\newblock {\em Natural capital: valuing the planet}; Yale University Press,
  2015.

\bibitem[Kumar(2012)]{kumar2012economics}
Kumar, P.
\newblock {\em The economics of ecosystems and biodiversity: ecological and
  economic foundations}; Routledge,  2012.

\bibitem[Sukhdev \em{et~al.}(2014)Sukhdev, Wittmer, and
  Miller]{sukhdev2014economics}
Sukhdev, P.; Wittmer, H.; Miller, D.
\newblock The economics of ecosystems and biodiversity (TEEB): challenges and
  responses.
\newblock {\em Nature in the balance: the economics of biodiversity} {\bf
  2014}, pp. 135--152.

\bibitem[Lubchenco(1998)]{lubchenco1998entering}
Lubchenco, J.
\newblock Entering the century of the environment: a new social contract for
  science.
\newblock {\em Science} {\bf 1998}, {\em 279},~491--497.

\bibitem[Wells and McShane(2004)]{wells2004integrating}
Wells, M.P.; McShane, T.O.
\newblock Integrating protected area management with local needs and
  aspirations.
\newblock {\em AMBIO: a Journal of the Human Environment} {\bf 2004}, {\em
  33},~513--519.

\bibitem[Parks and Tsioumani(2023)]{parks2023transforming}
Parks, L.; Tsioumani, E.
\newblock Transforming biodiversity governance? Indigenous peoples'
  contributions to the Convention on Biological Diversity.
\newblock {\em Biological Conservation} {\bf 2023}, {\em 280},~109933.

\bibitem[Brosius(2004)]{brosius2004indigenous}
Brosius, J.P.
\newblock Indigenous peoples and protected areas at the World Parks Congress.
\newblock {\em Conservation Biology} {\bf 2004}, {\em 18},~609--612.

\bibitem[Zurba \em{et~al.}(2019)Zurba, F.~Beazley, English, and
  Buchmann-Duck]{zurba2019indigenous}
Zurba, M.; F.~Beazley, K.; English, E.; Buchmann-Duck, J.
\newblock Indigenous protected and conserved areas (IPCAs), Aichi Target 11 and
  Canada’s Pathway to Target 1: Focusing conservation on reconciliation.
\newblock {\em Land} {\bf 2019}, {\em 8},~10.

\bibitem[IP(2021)]{ip2021wwf}
IP, C.
\newblock WWF GEF PROJECT DOCUMENT {\bf 2021}.

\bibitem[Ren \em{et~al.}(2015)Ren, He, Girshick, and Sun]{ren2015faster}
Ren, S.; He, K.; Girshick, R.; Sun, J.
\newblock Faster r-cnn: Towards real-time object detection with region proposal
  networks.
\newblock {\em Advances in neural information processing systems} {\bf 2015},
  {\em 28}.

\bibitem[con(11/04/2023)]{conservationai2023}
Conservation AI.
\newblock {\em https://www.conservationai.co.uk} {\bf 11/04/2023}.

\bibitem[Chalmers \em{et~al.}(2021)Chalmers, Fergus, Wich, and
  Longmore]{chalmers2021modelling}
Chalmers, C.; Fergus, P.; Wich, S.; Longmore, S.
\newblock Modelling Animal Biodiversity Using Acoustic Monitoring and Deep
  Learning.
\newblock In Proceedings of the 2021 International Joint Conference on Neural
  Networks (IJCNN). IEEE,  2021, pp. 1--7.

\bibitem[Lin \em{et~al.}(2014)Lin, Maire, Belongie, Hays, Perona, Ramanan,
  Doll{\'a}r, and Zitnick]{lin2014microsoft}
Lin, T.Y.; Maire, M.; Belongie, S.; Hays, J.; Perona, P.; Ramanan, D.;
  Doll{\'a}r, P.; Zitnick, C.L.
\newblock Microsoft coco: Common objects in context.
\newblock In Proceedings of the Computer Vision--ECCV 2014: 13th European
  Conference, Zurich, Switzerland, September 6-12, 2014, Proceedings, Part V
  13. Springer,  2014, pp. 740--755.

\bibitem[Welbourne \em{et~al.}(2016)Welbourne, Claridge, Paull, and
  Lambert]{welbourne2016passive}
Welbourne, D.J.; Claridge, A.W.; Paull, D.J.; Lambert, A.
\newblock How do passive infrared triggered camera traps operate and why does
  it matter? Breaking down common misconceptions.
\newblock {\em Remote Sensing in Ecology and Conservation} {\bf 2016}, {\em
  2},~77--83.

\bibitem[Gu \em{et~al.}(2018)Gu, Wang, Kuen, Ma, Shahroudy, Shuai, Liu, Wang,
  Wang, Cai, et~al.]{gu2018recent}
Gu, J.; Wang, Z.; Kuen, J.; Ma, L.; Shahroudy, A.; Shuai, B.; Liu, T.; Wang,
  X.; Wang, G.; Cai, J.;  et~al.
\newblock Recent advances in convolutional neural networks.
\newblock {\em Pattern recognition} {\bf 2018}, {\em 77},~354--377.

\bibitem[Ren \em{et~al.}(2016)Ren, He, Girshick, Zhang, and Sun]{ren2016object}
Ren, S.; He, K.; Girshick, R.; Zhang, X.; Sun, J.
\newblock Object detection networks on convolutional feature maps.
\newblock {\em IEEE transactions on pattern analysis and machine intelligence}
  {\bf 2016}, {\em 39},~1476--1481.

\bibitem[He \em{et~al.}(2016)He, Zhang, Ren, and Sun]{he2016deep}
He, K.; Zhang, X.; Ren, S.; Sun, J.
\newblock Deep residual learning for image recognition.
\newblock In Proceedings of the Proceedings of the IEEE conference on computer
  vision and pattern recognition,  2016, pp. 770--778.

\bibitem[Girshick(2015)]{girshick2015fast}
Girshick, R.
\newblock Fast r-cnn.
\newblock In Proceedings of the Proceedings of the IEEE international
  conference on computer vision,  2015, pp. 1440--1448.

\bibitem[Uijlings \em{et~al.}(2013)Uijlings, Van De~Sande, Gevers, and
  Smeulders]{uijlings2013selective}
Uijlings, J.R.; Van De~Sande, K.E.; Gevers, T.; Smeulders, A.W.
\newblock Selective search for object recognition.
\newblock {\em International journal of computer vision} {\bf 2013}, {\em
  104},~154--171.

\bibitem[Neubeck and Van~Gool(2006)]{neubeck2006efficient}
Neubeck, A.; Van~Gool, L.
\newblock Efficient non-maximum suppression.
\newblock In Proceedings of the 18th international conference on pattern
  recognition (ICPR'06). IEEE,  2006, Vol.~3, pp. 850--855.

\bibitem[Agarap(2018)]{agarap2018deep}
Agarap, A.F.
\newblock Deep learning using rectified linear units (relu).
\newblock {\em arXiv preprint arXiv:1803.08375} {\bf 2018}.

\bibitem[Rumelhart \em{et~al.}(1986)Rumelhart, Hinton, and
  Williams]{rumelhart1986learning}
Rumelhart, D.E.; Hinton, G.E.; Williams, R.J.
\newblock Learning representations by back-propagating errors.
\newblock {\em nature} {\bf 1986}, {\em 323},~533--536.

\bibitem[Robbins and Monro(1951)]{robbins1951stochastic}
Robbins, H.; Monro, S.
\newblock A stochastic approximation method.
\newblock {\em The annals of mathematical statistics} {\bf 1951}, pp. 400--407.

\bibitem[Pan and Yang(2010)]{pan2010survey}
Pan, S.J.; Yang, Q.
\newblock A survey on transfer learning.
\newblock {\em IEEE Transactions on knowledge and data engineering} {\bf 2010},
  {\em 22},~1345--1359.

\bibitem[Ying(2019)]{ying2019overview}
Ying, X.
\newblock An overview of overfitting and its solutions.
\newblock In Proceedings of the Journal of physics: Conference series. IOP
  Publishing,  2019, Vol. 1168, p. 022022.

\bibitem[Srivastava \em{et~al.}(2015)Srivastava, Greff, and
  Schmidhuber]{srivastava2015highway}
Srivastava, R.K.; Greff, K.; Schmidhuber, J.
\newblock Highway networks.
\newblock {\em arXiv preprint arXiv:1505.00387} {\bf 2015}.

\bibitem[Keckler \em{et~al.}(2011)Keckler, Dally, Khailany, Garland, and
  Glasco]{keckler2011gpus}
Keckler, S.W.; Dally, W.J.; Khailany, B.; Garland, M.; Glasco, D.
\newblock GPUs and the future of parallel computing.
\newblock {\em IEEE micro} {\bf 2011}, {\em 31},~7--17.

\bibitem[Goldsborough(2016)]{goldsborough2016tour}
Goldsborough, P.
\newblock A tour of tensorflow.
\newblock {\em arXiv preprint arXiv:1610.01178} {\bf 2016}.

\bibitem[Huang \em{et~al.}(2017)Huang, Rathod, Chow, Sun, Zhu, Fathi, and
  Lu]{huang2017tensorflow}
Huang, J.; Rathod, V.; Chow, D.; Sun, C.; Zhu, M.; Fathi, A.; Lu, Z.
\newblock Tensorflow object detection api.
\newblock {\em Code: github. com/tensorflow/models/tree/master/object
  detection} {\bf 2017}.

\bibitem[Kingma and Ba(2014)]{kingma2014adam}
Kingma, D.P.; Ba, J.
\newblock Adam: A method for stochastic optimization.
\newblock {\em arXiv preprint arXiv:1412.6980} {\bf 2014}.

\bibitem[Bottou(2012)]{bottou2012stochastic}
Bottou, L.
\newblock Stochastic gradient descent tricks.
\newblock {\em Neural Networks: Tricks of the Trade: Second Edition} {\bf
  2012}, pp. 421--436.

\bibitem[Sharma \em{et~al.}(2017)Sharma, Sharma, and
  Athaiya]{sharma2017activation}
Sharma, S.; Sharma, S.; Athaiya, A.
\newblock Activation functions in neural networks.
\newblock {\em Towards Data Sci} {\bf 2017}, {\em 6},~310--316.

\bibitem[Nair and Hinton(2010)]{nair2010rectified}
Nair, V.; Hinton, G.E.
\newblock Rectified linear units improve restricted boltzmann machines.
\newblock In Proceedings of the Icml,  2010.

\bibitem[Jahanshahi \em{et~al.}(2020)Jahanshahi, Sabzi, Lau, and
  Wong]{jahanshahi2020gpu}
Jahanshahi, A.; Sabzi, H.Z.; Lau, C.; Wong, D.
\newblock Gpu-nest: Characterizing energy efficiency of multi-gpu inference
  servers.
\newblock {\em IEEE Computer Architecture Letters} {\bf 2020}, {\em
  19},~139--142.

\bibitem[Caravaggi \em{et~al.}(2017)Caravaggi, Banks, Burton, Finlay, Haswell,
  Hayward, Rowcliffe, and Wood]{caravaggi2017review}
Caravaggi, A.; Banks, P.B.; Burton, A.C.; Finlay, C.M.; Haswell, P.M.; Hayward,
  M.W.; Rowcliffe, M.J.; Wood, M.D.
\newblock A review of camera trapping for conservation behaviour research.
\newblock {\em Remote Sensing in Ecology and Conservation} {\bf 2017}, {\em
  3},~109--122.

\bibitem[Postel(1982)]{postel1982simple}
Postel, J.
\newblock Simple mail transfer protocol.
\newblock Technical report,  1982.

\bibitem[Masse(2011)]{masse2011rest}
Masse, M.
\newblock {\em REST API design rulebook: designing consistent RESTful web
  service interfaces}; " O'Reilly Media, Inc.",  2011.

\bibitem[pay()]{paypaldeveloper}
Build a payment solution that's right for you with PayPal for developers.

\bibitem[Padilla \em{et~al.}(2020)Padilla, Netto, and
  Da~Silva]{padilla2020survey}
Padilla, R.; Netto, S.L.; Da~Silva, E.A.
\newblock A survey on performance metrics for object-detection algorithms.
\newblock In Proceedings of the 2020 international conference on systems,
  signals and image processing (IWSSIP). IEEE,  2020, pp. 237--242.

\bibitem[Palencia \em{et~al.}(2022)Palencia, Vicente, Soriguer, and
  Acevedo]{palencia2022towards}
Palencia, P.; Vicente, J.; Soriguer, R.C.; Acevedo, P.
\newblock Towards a best-practices guide for camera trapping: assessing
  differences among camera trap models and settings under field conditions.
\newblock {\em Journal of Zoology} {\bf 2022}, {\em 316},~197--208.

\bibitem[Duffy and St~John(2013)]{duffy2013poverty}
Duffy, R.; St~John, F.
\newblock Poverty, Poaching and Trafficking: What are the links? {\bf 2013}.

\bibitem[Ayompe \em{et~al.}(2021)Ayompe, Nkongho, Masso, and
  Egoh]{ayompe2021does}
Ayompe, L.M.; Nkongho, R.N.; Masso, C.; Egoh, B.N.
\newblock Does investment in palm oil trade alleviate smallholders from poverty
  in Africa? Investigating profitability from a biodiversity hotspot, Cameroon.
\newblock {\em PloS one} {\bf 2021}, {\em 16},~e0256498.

\bibitem[Costanza(2020)]{costanza2020valuing}
Costanza, R.
\newblock Valuing natural capital and ecosystem services toward the goals of
  efficiency, fairness, and sustainability.
\newblock {\em Ecosystem Services} {\bf 2020}, {\em 43},~101096.

\bibitem[Talukdar \em{et~al.}(2018)Talukdar, Singh, and
  Choudhury]{talukdar2018conservation}
Talukdar, N.R.; Singh, B.; Choudhury, P.
\newblock Conservation status of some endangered mammals in Barak Valley,
  Northeast India.
\newblock {\em Journal of Asia-Pacific Biodiversity} {\bf 2018}, {\em
  11},~167--172.

\bibitem[Courchamp \em{et~al.}(2006)Courchamp, Angulo, Rivalan, Hall, Signoret,
  Bull, and Meinard]{courchamp2006rarity}
Courchamp, F.; Angulo, E.; Rivalan, P.; Hall, R.J.; Signoret, L.; Bull, L.;
  Meinard, Y.
\newblock Rarity value and species extinction: the anthropogenic Allee effect.
\newblock {\em PLoS biology} {\bf 2006}, {\em 4},~e415.

\bibitem[Waldram \em{et~al.}(2008)Waldram, Bond, and
  Stock]{waldram2008ecological}
Waldram, M.S.; Bond, W.J.; Stock, W.D.
\newblock Ecological engineering by a mega-grazer: white rhino impacts on a
  South African savanna.
\newblock {\em Ecosystems} {\bf 2008}, {\em 11},~101--112.

\bibitem[Berkes(2017)]{berkes2017sacred}
Berkes, F.
\newblock {\em Sacred ecology}; Routledge,  2017.

\bibitem[Pierotti(2010)]{pierotti2010indigenous}
Pierotti, R.
\newblock {\em Indigenous knowledge, ecology, and evolutionary biology};
  Routledge,  2010.

\bibitem[Russakovsky \em{et~al.}(2015)Russakovsky, Deng, Su, Krause, Satheesh,
  Ma, Huang, Karpathy, Khosla, Bernstein, et~al.]{russakovsky2015imagenet}
Russakovsky, O.; Deng, J.; Su, H.; Krause, J.; Satheesh, S.; Ma, S.; Huang, Z.;
  Karpathy, A.; Khosla, A.; Bernstein, M.;  et~al.
\newblock Imagenet large scale visual recognition challenge.
\newblock {\em International journal of computer vision} {\bf 2015}, {\em
  115},~211--252.

\bibitem[Swanson \em{et~al.}(2015)Swanson, Kosmala, Lintott, Simpson, Smith,
  and Packer]{swanson2015snapshot}
Swanson, A.; Kosmala, M.; Lintott, C.; Simpson, R.; Smith, A.; Packer, C.
\newblock Snapshot Serengeti, high-frequency annotated camera trap images of 40
  mammalian species in an African savanna.
\newblock {\em Scientific data} {\bf 2015}, {\em 2},~1--14.

\bibitem[Tabak \em{et~al.}(2019)Tabak, Norouzzadeh, Wolfson, Sweeney,
  VerCauteren, Snow, Halseth, Di~Salvo, Lewis, White, et~al.]{tabak2019machine}
Tabak, M.A.; Norouzzadeh, M.S.; Wolfson, D.W.; Sweeney, S.J.; VerCauteren,
  K.C.; Snow, N.P.; Halseth, J.M.; Di~Salvo, P.A.; Lewis, J.S.; White, M.D.;
  et~al.
\newblock Machine learning to classify animal species in camera trap images:
  Applications in ecology.
\newblock {\em Methods in Ecology and Evolution} {\bf 2019}, {\em
  10},~585--590.

\bibitem[Willi \em{et~al.}(2019)Willi, Pitman, Cardoso, Locke, Swanson, Boyer,
  Veldthuis, and Fortson]{willi2019identifying}
Willi, M.; Pitman, R.T.; Cardoso, A.W.; Locke, C.; Swanson, A.; Boyer, A.;
  Veldthuis, M.; Fortson, L.
\newblock Identifying animal species in camera trap images using deep learning
  and citizen science.
\newblock {\em Methods in Ecology and Evolution} {\bf 2019}, {\em 10},~80--91.

\bibitem[Yousif \em{et~al.}(2017)Yousif, Yuan, Kays, and He]{yousif2017fast}
Yousif, H.; Yuan, J.; Kays, R.; He, Z.
\newblock Fast human-animal detection from highly cluttered camera-trap images
  using joint background modeling and deep learning classification.
\newblock In Proceedings of the 2017 IEEE international symposium on circuits
  and systems (ISCAS). IEEE,  2017, pp. 1--4.

\bibitem[Norouzzadeh \em{et~al.}(2018)Norouzzadeh, Nguyen, Kosmala, Swanson,
  Palmer, Packer, and Clune]{norouzzadeh2018automatically}
Norouzzadeh, M.S.; Nguyen, A.; Kosmala, M.; Swanson, A.; Palmer, M.S.; Packer,
  C.; Clune, J.
\newblock Automatically identifying, counting, and describing wild animals in
  camera-trap images with deep learning.
\newblock {\em Proceedings of the National Academy of Sciences} {\bf 2018},
  {\em 115},~E5716--E5725.

\bibitem[Norouzzadeh \em{et~al.}(2021)Norouzzadeh, Morris, Beery, Joshi, Jojic,
  and Clune]{norouzzadeh2021deep}
Norouzzadeh, M.S.; Morris, D.; Beery, S.; Joshi, N.; Jojic, N.; Clune, J.
\newblock A deep active learning system for species identification and counting
  in camera trap images.
\newblock {\em Methods in ecology and evolution} {\bf 2021}, {\em
  12},~150--161.

\bibitem[Villa \em{et~al.}(2017)Villa, Salazar, and Vargas]{villa2017towards}
Villa, A.G.; Salazar, A.; Vargas, F.
\newblock Towards automatic wild animal monitoring: Identification of animal
  species in camera-trap images using very deep convolutional neural networks.
\newblock {\em Ecological informatics} {\bf 2017}, {\em 41},~24--32.

\bibitem[Witter \em{et~al.}(2015)Witter, Marion~Suiseeya, Gruby, Hitchner,
  Maclin, Bourque, and Brosius]{witter2015moments}
Witter, R.; Marion~Suiseeya, K.R.; Gruby, R.L.; Hitchner, S.; Maclin, E.M.;
  Bourque, M.; Brosius, J.P.
\newblock Moments of influence in global environmental governance.
\newblock {\em Environmental Politics} {\bf 2015}, {\em 24},~894--912.

\bibitem[Sachedina(2010)]{sachedina2010disconnected}
Sachedina, H.T.
\newblock Disconnected nature: the scaling up of African Wildlife Foundation
  and its impacts on biodiversity conservation and local livelihoods.
\newblock {\em Antipode} {\bf 2010}, {\em 42},~603--623.

\bibitem[Zheng \em{et~al.}(2018)Zheng, Xie, Dai, Chen, and
  Wang]{zheng2018blockchain}
Zheng, Z.; Xie, S.; Dai, H.N.; Chen, X.; Wang, H.
\newblock Blockchain challenges and opportunities: A survey.
\newblock {\em International journal of web and grid services} {\bf 2018}, {\em
  14},~352--375.

\bibitem[Dowie(2011)]{dowie2011conservation}
Dowie, M.
\newblock {\em Conservation refugees: the hundred-year conflict between global
  conservation and native peoples}; MIT press,  2011.

\end{thebibliography}

%


\end{adjustwidth}
\end{document}